\documentclass[journal,compsoc]{IEEEtran}
\usepackage{graphicx,float,amssymb,multirow,pifont,amsmath,amsfonts,scalerel,makecell}
\usepackage{bm,xcolor,enumitem,booktabs,arydshln,footnote}
\usepackage{array}
\usepackage[caption=false,font=normalsize,labelfont=sf,textfont=sf]{subfig}
\usepackage{textcomp}
\usepackage{stfloats}
\usepackage{url}
\usepackage{verbatim}
\usepackage{color}
\usepackage{amsthm,amsmath,algpseudocode,graphicx}
\usepackage{threeparttable}
\usepackage[linesnumbered,ruled,vlined]{algorithm2e}
\usepackage[subfigure,titles]{tocloft}
\usepackage[colorlinks, linkcolor=blue, citecolor=blue]{hyperref}
\usepackage{cite}
\usepackage[a4paper,top=1.4cm,bottom=1.4cm,left=1.4cm,right=1.4cm]{geometry}
\graphicspath{{figures/}}
\usepackage{balance}
\usepackage{placeins}
\begin{document}
\title{Relaxing Cross-Modal Correspondence for Multimodal Remote Sensing Pretraining}
\author{Xianping Ma,
		Yu Zhang,
		Mingzhe Jiang,
		Wanxin Xie,
		Qing Zhu,
		Nan Xu,
		Xiaokang Zhang,
		and Yuanxin Ye$^{\ast}$
		\thanks{Xianping Ma, Yu Zhang, Mingzhe Jiang, Wanxin Xie, Qing Zhu, and Yuanxin Ye are with the Faculty of Geosciences and Engineering, Southwest Jiaotong University, Chengdu 611756, China (e-mail: mxp@swjtu.edu.cn; yeyuanxin@swjtu.edu.cn).}
		\thanks{Nan Xu is with the School of Architecture and Urban Planning, Shenzhen University, Shenzhen 518060, China.}
        \thanks{Xiaokang~Zhang is with the School of Artificial Intelligence, Wuhan University, Wuhan 430072, China.}}

\markboth{Preprint}%
{Relaxing Cross-Modal Correspondence for Multimodal Remote Sensing Pretraining}

\maketitle

\begin{abstract}
Multimodal remote sensing pretraining commonly relies on exact instance-level correspondence across modalities, implicitly treating strictly paired observations as a prerequisite for learning transferable representations. In this work, we revisit this assumption by relaxing cross-modal correspondence from the instance level to the semantic level. We introduce Semantic Pairing (\textbf{SemPair}), a pairing-relaxed pretraining strategy that replaces exact instance pairs with class-consistent cross-instance pairs. To characterize the effect of correspondence relaxation, we formulate a latent-factor model that separates cross-modal dependence into shared semantic structure and instance-specific correlation, showing that SemPair preserves the former while marginalizing instance-specific cross-modal correlations in expectation. We further introduce Mixed Pairing (\textbf{MixPair}) as a controlled interpolation between semantic and exact pairing to study correspondence granularity along a continuous spectrum. Extensive experiments consistently show that SemPair achieves transfer performance comparable to exact pairing (ExaPair), while MixPair reveals a non-monotonic relationship between downstream transfer and the instance-pairing ratio. These results demonstrate that exact instance correspondence is not a prerequisite for effective multimodal remote sensing pretraining and establish cross-modal correspondence granularity as an important design dimension for transferable multimodal representation learning. More broadly, this finding suggests a practical route toward scaling multimodal pretraining with reduced dependence on costly instance-aligned data.
\end{abstract}


\section{Introduction}
Earth observation increasingly relies on multiple sensing modalities that provide complementary spectral, geometric, structural, and scattering information \cite{zhu2017deep,ma2019deep}. Remote sensing research is also moving from task-specific learning toward reusable pretraining and transfer, making general representation learning an increasingly important problem \cite{lu2025vfmrs}. Recent multimodal foundation models have shown that large-scale heterogeneous observations can support more general Earth observation representations \cite{guo2024skysense}. This trend has continued toward larger multimodal corpora and broader downstream applicability \cite{wu2025skysense}. A common assumption in these systems, however, is that observations from different modalities should correspond to the same physical instance or spatial location during pretraining.

Exact instance pairing provides a clear cross-modal relation, but the information captured by such correspondence is not necessarily identical to the information required for downstream transfer. The goal of pretraining is to learn representations that remain useful when tasks, regions, and sensing conditions change, rather than to reproduce the source domain pairing relation itself \cite{bengio2013representation,jing2021ssl}. Paired observations contain both semantic information that can recur across different instances and details that are specific to individual samples. Recent multimodal representation learning has also shown that shared and modality-specific information can play different roles in learned representations \cite{wang2024decur}. These observations raise a central question: \textbf{Is exact instance correspondence necessary for learning transferable multimodal representations?}

Remote sensing makes this question particularly important. Large-scale Earth observation data vary substantially in sensor type, spatial resolution, acquisition time, geographic coverage, and available supervision. Recent pretraining efforts have increasingly focused on learning from heterogeneous sensors, resolutions, and spatial scales \cite{astruc2025anysat}. Large multimodal models are also beginning to integrate a wider range of Earth observation signals within a common representation space \cite{jakubik2025terramind}. At the data level, however, multimodal supervision is still commonly constructed from observations matched by spatial location and acquisition time \cite{nedungadi2024mmearth}. Requiring all samples to form strict one-to-one multimodal pairs can nevertheless restrict the amount and diversity of data that can be readily organized for pretraining. 

Existing multimodal learning mainly studies how heterogeneous information should be represented, aligned, fused, and transferred across modalities \cite{ramachandram2017deep,baltrusaitis2019multimodal}. Remote sensing research has similarly explored complementary information across heterogeneous observations and knowledge transfer between sensing domains \cite{audebert2018beyondrgb,hong2020xmodal}. More recent multisensor and multimodal foundation models have extended this direction to broader sensor combinations and unified representation learning \cite{guo2024skysense,zhang2025skysensev2}. Despite this progress, the correspondence relation used during pretraining is still largely determined by the available data rather than treated as a controllable learning variable. Whether exact instance correspondence is necessary, and how transferability changes when it is relaxed, therefore remain insufficiently understood. 

\begin{figure*}[!t]
\centering
\includegraphics[width=0.95\linewidth]{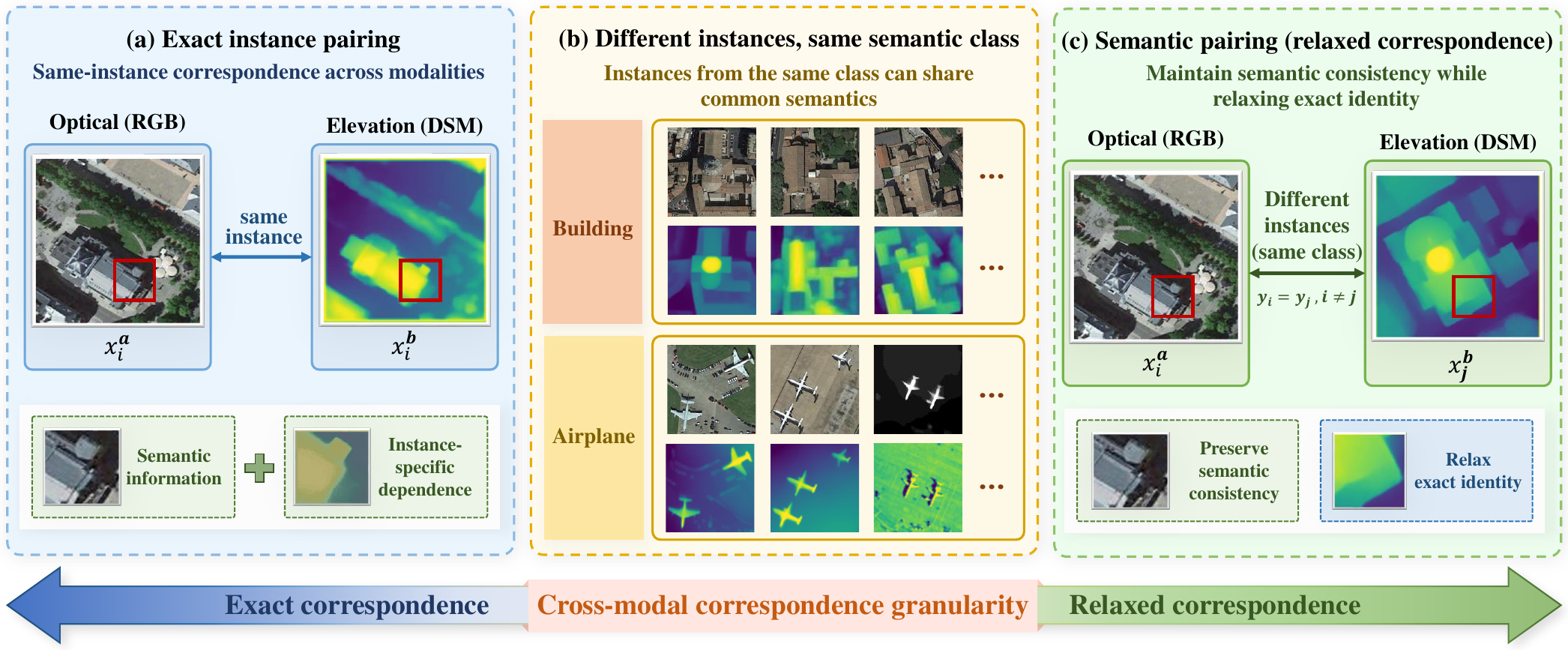}
\caption{Motivation for relaxing exact cross-modal correspondence in multimodal pretraining. ExaPair enforces same-instance correspondence, coupling semantic consistency with instance-specific dependence. Different instances from the same semantic class can still share common semantic structure despite variations in appearance and geometry. This motivates SemPair, which preserves semantic consistency while relaxing exact instance identity.}
\label{fig:pairing_space}
\end{figure*}
This study addresses this gap by considering \emph{cross-modal correspondence granularity} as an explicit design dimension of multimodal pretraining. Instead of treating exact pairing (ExaPair) as a fixed requirement, we explore how much correspondence should be preserved between modalities to learn transferable representations. This view places different levels of cross-modal correspondence within a common space, allowing the effect of correspondence level to be studied independently of changes in model architecture or training objective. Figure~\ref{fig:pairing_space} illustrates this perspective by contrasting exact instance pairing with semantic pairing, where semantic consistency is preserved while exact instance identity is relaxed.

We investigate this problem through a controlled pairing-relaxed pretraining framework, using Semantic Pairing (\textbf{SemPair}) and Mixed Pairing (\textbf{MixPair}) to study different levels of cross-modal correspondence. We further combine theoretical analysis with extensive transfer experiments to examine the role of correspondence granularity in multimodal pretraining. The main contributions of this work are summarized as follows:
\begin{itemize}
\item We formulate cross-modal correspondence granularity as an explicit design dimension for multimodal remote sensing pretraining. Instead of assuming exact instance-aligned pairs as a necessary form of supervision, we investigate how much cross-modal correspondence should be preserved to learn transferable multimodal representations.
\item We introduce SemPair, a pairing-relaxed pretraining strategy that replaces exact instance correspondence with class-consistent cross-instance correspondence, without modifying the network architecture, training objective, or optimization budget. We further generalize ExaPair and SemPair into a continuous correspondence spectrum through MixPair, and provide a latent-factor analysis showing how correspondence granularity regulates the balance between shared semantic structure and instance-specific dependence.
\item Extensive transfer experiments show that relaxing exact instance correspondence causes little loss in downstream performance. SemPair remains highly competitive with ExaPair across four multimodal semantic segmentation benchmarks, while the continuous pairing-ratio study further reveals that transferability does not improve monotonically with the instance-pairing ratio. These results demonstrate that ExaPair is neither indispensable nor consistently optimal for transferable multimodal pretraining.
\item We systematically evaluate pairing-relaxed pretraining across two pretraining domains, four downstream multimodal segmentation benchmarks, different supervision and adaptation regimes, and multiple model scales. The results consistently demonstrate its effectiveness across diverse source-target conditions, establishing the robustness and generality of the proposed correspondence-relaxation paradigm.
\end{itemize}
This work provides a practical perspective for constructing larger multimodal pretraining datasets. By reducing the dependence on strictly paired observations, it offers a more flexible way to organize and utilize heterogeneous remote sensing data for pretraining. The remainder of this paper is organized as follows. Section~\ref{sec:related} reviews related work. Section~\ref{sec:metho} presents the proposed pairing-relaxed pretraining framework and its theoretical analysis. Section~\ref{sec:exper} provides the experimental evaluation, analyses the effect of correspondence granularity under different transfer settings, and discusses the scope and limitations of the study. Finally, Section~\ref{sec:conclu} concludes the paper.

\section{Related Work}\label{sec:related}
\subsection{Multimodal Representation Learning in Remote Sensing}
Multimodal remote sensing combines observations from different sensing mechanisms to exploit complementary information \cite{khaleghi2013multisensor,zhang2021multimodalseg,li2022multimodalreview}. Early deep-learning approaches mainly focused on how heterogeneous information should be represented and fused. Optical-elevation fusion demonstrated the value of combining appearance and geometric cues for very-high-resolution scene understanding \cite{audebert2018beyondrgb}. Subsequent studies explored joint representation learning across heterogeneous observations \cite{hong2021more}, knowledge transfer between sensing domains \cite{hong2020xmodal}, and the decomposition of shared and modality-specific information \cite{hong2021s2fl}. Manifold-based alignment has also been used to relate observations with substantially different sensing characteristics \cite{hu2019mima}. These studies established the benefit of complementary sensing, but generally operated on observations that were already matched at the instance or spatial level.

Recent architectures have further strengthened cross-modal interaction through more structured feature exchange. CMGFNet \cite{hosseinpour2022cmgfnet} introduces gated interaction between optical and elevation features, while Transformer-based FTransUNet \cite{ma2024ftransunet} captures cross-modal dependencies across multiple representation levels. Feature compensation and reconstruction have also been introduced to improve the use of complementary information under modality differences \cite{wang2025cfcgnet}. Large pretrained encoders can further be adapted to optical-DSM segmentation through multimodal fine-tuning \cite{ma2025mfnet,qi2026multimoe}. Although these methods differ considerably in architecture and fusion mechanism, their main focus is how corresponding multimodal observations should interact. The correspondence relation between the inputs is usually inherited from the dataset and remains outside the model design.

\subsection{Multimodal Remote Sensing Pretraining}
A parallel line of research aims to learn reusable representations from large-scale remote sensing data. CMID \cite{muhtar2023cmid} combines contrastive learning with masked image modeling, while RingMo \cite{sun2023ringmo} develops large-scale masked pretraining for remote sensing imagery. SpectralGPT \cite{hong2024spectralgpt} further extends foundation modeling to spatial-spectral representations. These studies demonstrate the effectiveness of pretraining for transferring knowledge across downstream tasks, but primarily focus on improving representation objectives or expanding the scale and diversity of the pretraining corpus.

Multisensor pretraining further exploits complementary observations across sensing modalities. SSL4EO-S12 \cite{wang2023ssl4eo} provides large-scale SAR-optical observations for self-supervised representation learning, and multimodal self-supervision has been used to reduce dependence on downstream annotations \cite{xue2025mmssl}. More recent foundation models extend this direction toward increasingly heterogeneous sensors, resolutions, and acquisition conditions \cite{guo2024skysense,astruc2025anysat}. Large-scale multimodal frameworks further seek unified representations across diverse Earth observation signals \cite{jakubik2025terramind,zhang2025skysensev2}. These developments indicate a clear trend toward broader multimodal pretraining and increasingly heterogeneous data sources.

Despite this progress, cross-modal supervision in such frameworks is still commonly constructed from observations that are spatially or temporally associated. When exact multimodal pairs are available, they are typically used directly as the basic correspondence unit for pretraining. As a result, improvements in architecture, objective, data scale, and sensor diversity have received substantial attention, whereas the correspondence relation itself has rarely been isolated as an independent variable. It therefore remains difficult to determine whether the benefits of multimodal pretraining arise from semantic information shared across modalities, from exact instance-level correspondence, or from both. This distinction becomes increasingly important when pretraining is expected to scale to heterogeneous data for which strict one-to-one pairing may be costly or unavailable.

\subsection{Relaxed Cross-Modal Correspondence Learning}
Broader multimodal learning has begun to consider incomplete, unaligned, and pair-free data, although these settings span substantially different modality combinations and learning objectives \cite{zong2025ssmlsurvey}. For image-image learning, several recent studies are more closely related to the problem considered here. MMGL \cite{wan2023mmgl} introduces self-supervised RGB-infrared pretraining with cross-modal cycle consistency for person re-identification under unpaired settings, while MCL \cite{yao2025unpaired} constructs pseudo cross-modal identity relations to support visible-infrared ReID without paired supervision. More recently, APTP and UPTP in \emph{Beyond Strict Pairing} \cite{deng2026beyondpairing} explicitly challenge strictly paired infrared-visible image fusion and show that arbitrary or unpaired training can substantially relax the need for rigidly aligned inputs. These studies demonstrate that exact image-level pairing is not always indispensable, but they mainly address task-specific alignment, identity association, or fusion objectives rather than transferable multimodal pretraining.

The necessity of exact correspondence depends strongly on the learning objective. Multimodal registration and image matching explicitly require geometric or spatial correspondence, and relaxing this relation would change the task definition \cite{geng2025matching}. Semantic representation learning is different, because useful information can be shared across instances even when their local appearance, geometry, or identity is not exactly matched. Existing image-image studies have shown that strict pairing can be relaxed under specific tasks, but the broader role of cross-modal correspondence in learning transferable representations remains insufficiently understood. In particular, it is still unclear how much correspondence should be preserved between modalities and how different correspondence levels affect downstream transfer. This gap motivates our study of \emph{cross-modal correspondence granularity} in multimodal remote sensing pretraining.

\begin{figure*}[!t]
\centering
\includegraphics[width=0.95\linewidth]{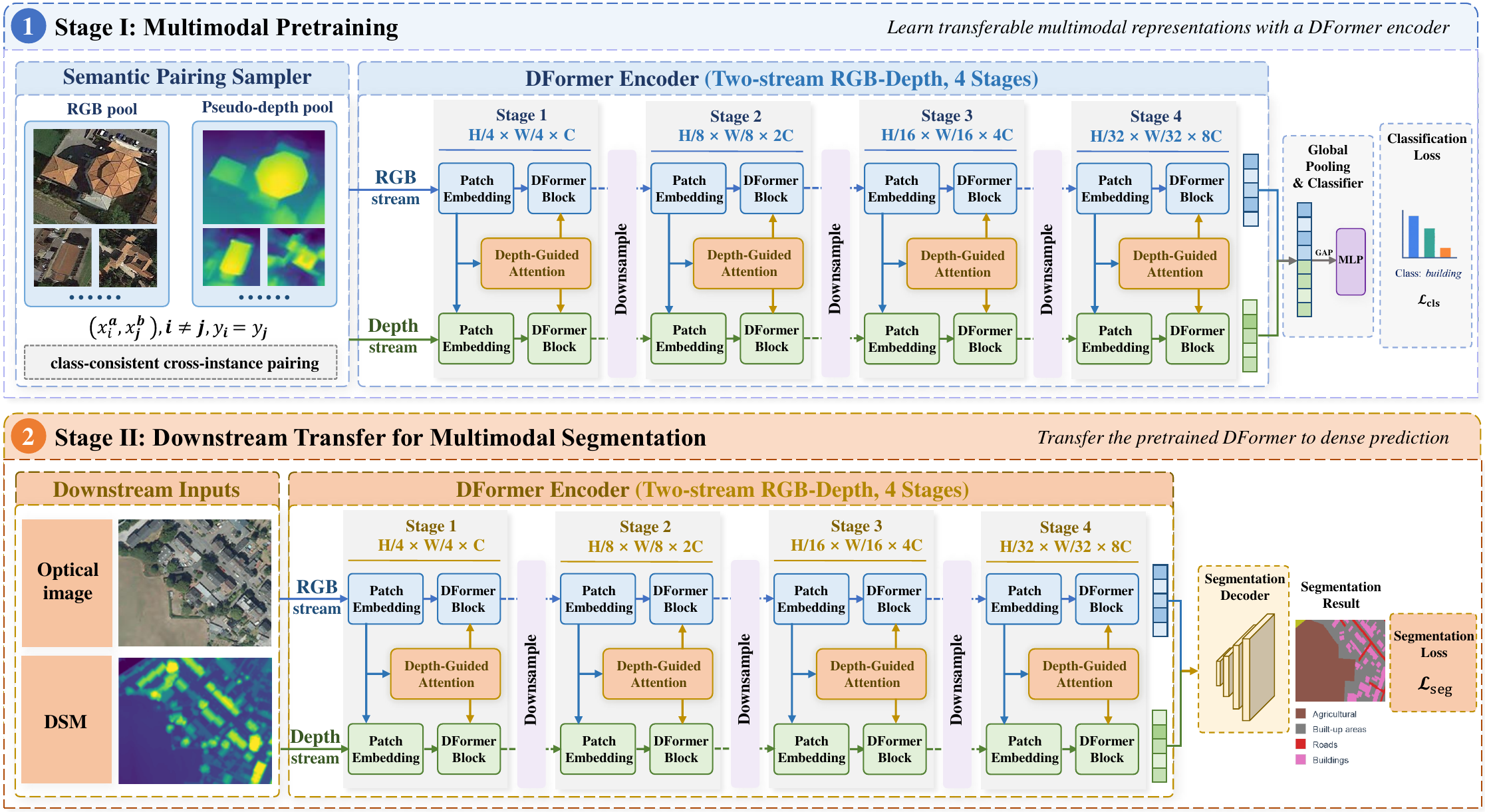}
\caption{Overview of pairing-relaxed multimodal pretraining and downstream transfer. Modality-$a$ anchors are paired with modality-$b$ samples according to the correspondence strategy. Only the cross-modal association varies among ExaPair, SemPair, and MixPair, while the multimodal encoder, pretraining objective, optimization budget, and downstream transfer protocol remain fixed.}
\label{fig:framework_overview}
\end{figure*}

\section{Methodology}\label{sec:metho}
\subsection{Problem Formulation and Overview}\label{sec:overview}
We formulate cross-modal correspondence granularity as a controllable factor in multimodal pretraining. Specifically, we vary the correspondence between the two input modalities while keeping the multimodal architecture and learning objective unchanged. Let the labeled pretraining set be
\begin{equation}
\mathcal{D}=\left\{(\mathbf{x}_{i}^{a},\mathbf{x}_{i}^{b},y_i)\right\}_{i=1}^{N},
\label{eq:dataset}
\end{equation}
where $\mathbf{x}_{i}^{a}$ and $\mathbf{x}_{i}^{b}$ are observations of two modalities associated with instance $i$, and $y_i\in\{1,\ldots,C\}$ is the corresponding class label. In our controlled pretraining experiments, modality $a$ is RGB and modality $b$ is pseudo-depth, while downstream transfer uses optical imagery and DSM. The source corpus contains exact pairs, allowing cross-modal correspondence to be deliberately changed without introducing additional samples or altering the semantic labels and training budget. Importantly, SemPair does not use the known exact associations during pretraining.

For compactness, we denote the fused prediction of the multimodal network by
\begin{equation}
\widehat{\mathbf{p}}_{ij}=g_{\phi}\left(f_{\theta}(\mathbf{x}_{i}^{a},\mathbf{x}_{j}^{b})\right),
\label{eq:network}
\end{equation}
where $f_{\theta}$ denotes the jointly interactive multimodal encoder and $g_{\phi}$ denotes the prediction head. Branch-specific predictions involved in the pretraining objective are omitted here for clarity. Conventional ExaPair fixes $j=i$, whereas our intervention changes only how the partner index $j$ is selected. Cross-modal correspondence is therefore treated as a property of the training distribution rather than a hard architectural requirement.

After pretraining, the classification head $g_{\phi}$ is discarded and the pretrained multimodal encoder $f_{\theta}$ is transferred to downstream multimodal semantic segmentation. The downstream inputs consist of naturally corresponding optical imagery and DSM, and a segmentation decoder is attached to the transferred encoder for dense prediction. No re-pairing is performed at this stage, as correspondence relaxation is applied only during pretraining.

The overall pipeline is shown in Figure~\ref{fig:framework_overview}. ExaPair, SemPair, and MixPair share the same network architecture, pretraining objective, and downstream transfer protocol. The only variable during pretraining is the pairing sampler, which controls the association between the two modality streams.

\subsection{Pairing-Relaxed Multimodal Pretraining}\label{sec:pair_construction}
\subsubsection{ExaPair and SemPair}
For each semantic class $c$, we first define the class-indexed training set
\begin{equation}
\mathcal{I}_{c}=\{k\mid y_k=c\}.
\label{eq:class_index}
\end{equation}

For an anchor instance $i$, its same-class cross-instance candidate set is
\begin{equation}
\mathcal{C}(i)=\{j\mid y_j=y_i,\;j\neq i\}.
\label{eq:candidate_set}
\end{equation}

ExaPair sets $j=i$ and therefore preserves exact instance correspondence across modalities. SemPair instead samples the modality-$b$ partner according to
\begin{equation}
p_{\mathrm{S}}(j\mid i)=\frac{\mathbb{I}[j\in\mathcal{C}(i)]}{|\mathcal{C}(i)|},
\qquad
j\sim p_{\mathrm{S}}(\cdot\mid i),
\label{eq:semantic_partner}
\end{equation}
where $\mathbb{I}[\cdot]$ denotes the indicator function. Thus, all candidates in $\mathcal{C}(i)$ are sampled with equal probability, ensuring that the two inputs share the same semantic label while corresponding to different instances. If $\mathcal{C}(i)$ is empty, SemPair falls back to the exact pair for that anchor. This case does not occur in the two pretraining datasets used in our experiments because every class contains multiple training samples. All candidate sets are constructed exclusively from the training partition, and the modality-$a$ anchor $\mathbf{x}_{i}^{a}$ remains unchanged.

SemPair is therefore different from random unpairing. An arbitrary cross-class partner could make the multimodal input semantically inconsistent with its supervision label. It is also different from geometric perturbation or modality dropout: both inputs remain complete and valid observations, while only their exact instance correspondence is relaxed. In our implementation, the pseudo-depth input for partner $j$ is the precomputed depth map belonging to sample $j$ and is never regenerated from, or copied from, anchor $i$.

We denote the training distributions induced by ExaPair and SemPair as $\mathbb{P}_{\mathrm{E}}$ and $\mathbb{P}_{\mathrm{S}}$, respectively. They preserve the same anchor and semantic-label marginals and differ only in the conditional association of modality-$b$ given the anchor.

\subsubsection{MixPair and Correspondence Ratio}
MixPair connects ExaPair and SemPair through a continuous correspondence spectrum. We define $\rho\in[0,1]$ as the \emph{instance-pairing ratio}, where $\rho=1$ corresponds to ExaPair and $\rho=0$ corresponds to SemPair.

For a requested ratio $\rho$, let $N_{\mathrm{E}}=\lfloor\rho N\rfloor$ denote the number of anchors assigned to ExaPair in each epoch. The realized ratio $N_{\mathrm{E}}/N$ differs from the requested value by less than $1/N$. For simplicity, we use $\rho$ to denote the pairing ratio hereafter. At the beginning of epoch $e$, let $\pi_e$ be a random permutation of all $N$ training indices and define
\begin{equation}
\mathcal{I}_{e}^{\mathrm{E}}=\{\pi_e(1),\ldots,\pi_e(N_{\mathrm{E}})\}.
\label{eq:epoch_exact_set}
\end{equation}

The modality-$b$ partner for anchor $i$ is then selected as
\begin{equation}
j_e(i)=
\begin{cases}
i, & i\in\mathcal{I}_{e}^{\mathrm{E}},\\[1mm]
j,\quad j\sim p_{\mathrm{S}}(\cdot\mid i),
& i\notin\mathcal{I}_{e}^{\mathrm{E}}\ \text{and}\ |\mathcal{C}(i)|>0,\\[1mm]
i, & |\mathcal{C}(i)|=0.
\end{cases}
\label{eq:partner_rule}
\end{equation}

The global permutation is independent of class identity. MixPair therefore does not impose a class-stratified allocation between ExaPair and SemPair, as it controls only the global epoch-level proportion. The assignment is regenerated at every epoch, allowing the same anchor to experience different correspondence modes during training. This epoch-wise reshuffling prevents the pairing mode from being permanently tied to particular samples and makes $\rho$ reflect the overall exposure to exact instance correspondence during training rather than a fixed partition of the dataset. Unless otherwise stated, MixPair uses $\rho=0.5$.

\begin{figure*}[!t]
\centering
\includegraphics[width=0.8\linewidth]{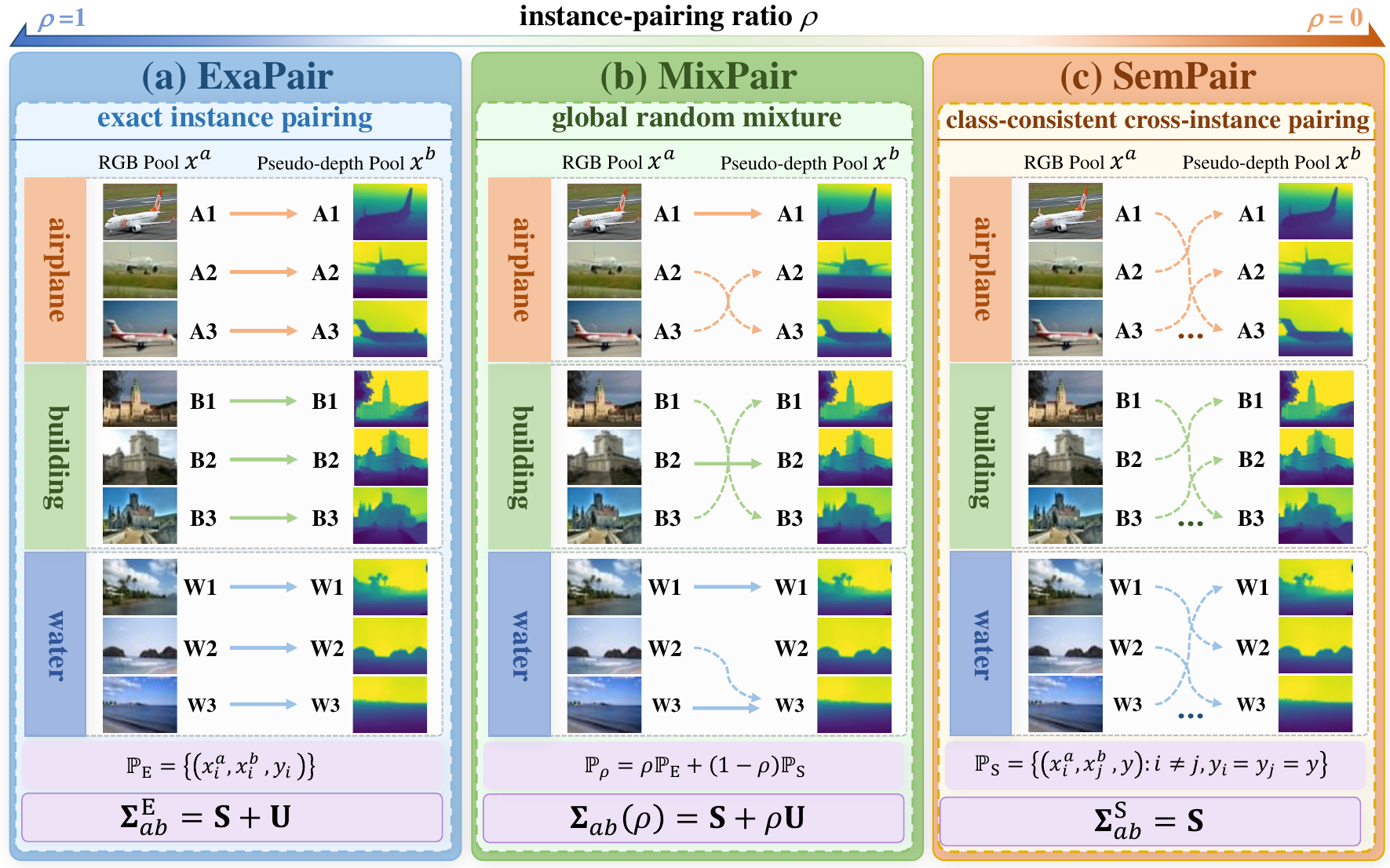}
\caption{ExaPair, SemPair, and MixPair for controlling cross-modal correspondence during pretraining. ExaPair preserves exact instance correspondence. SemPair keeps the modality-$a$ anchor fixed and replaces modality $b$ with a different sample from the same semantic class. MixPair assigns a fraction $\rho$ of training anchors to ExaPair and the remaining fraction $1-\rho$ to SemPair, with the assignment reshuffled globally at every epoch.}
\label{fig:pairing_strategies}
\end{figure*}

Figure~\ref{fig:pairing_strategies} summarizes the three pairing strategies. ExaPair and SemPair form the two endpoints of the correspondence spectrum, while MixPair provides intermediate correspondence levels through $\rho$. Because anchor sampling is unchanged and $y_j=y_i$ under SemPair, varying $\rho$ does not alter the number of anchors, the anchor-label distribution, or the number of optimizer steps. It changes only which modality-$b$ observation accompanies each modality-$a$ anchor.

\subsubsection{Pretraining Procedure and Controlled Comparison}
Algorithm~\ref{alg:pairing_relaxed} summarizes the training procedure. Class-indexed candidate sets are constructed once from the training partition. At each epoch, MixPair first shuffles all training anchors and assigns the configured fraction to ExaPair. Mini-batches are then assembled according to Eq.~\eqref{eq:partner_rule} and optimized using the original pretraining objective.

\begin{algorithm*}[t]
\caption{Multimodal Pretraining with Controlled Cross-Modal Correspondence}
\label{alg:pairing_relaxed}
\DontPrintSemicolon

\KwIn{Pretraining set $\mathcal{D}$, instance-pairing ratio $\rho$, network $(f_{\theta},g_{\phi})$}
\KwOut{Pretrained multimodal encoder $f_{\theta^{*}}$}

Build class-indexed sets $\{\mathcal{I}_{c}\}_{c=1}^{C}$ from the training split\;
Set $N_{\mathrm{E}}\leftarrow\lfloor\rho N\rfloor$
\tcp*[r]{$\rho=1$: ExaPair; $\rho=0$: SemPair; otherwise: MixPair}
\ForEach{epoch $e$}{
    Randomly permute all training indices and construct
    $\mathcal{I}_{e}^{\mathrm{E}}$ with $N_{\mathrm{E}}$ anchors\;
    \ForEach{mini-batch $\mathcal{B}$}{
        \ForEach{$i\in\mathcal{B}$}{
            \eIf{$i\in\mathcal{I}_{e}^{\mathrm{E}}$}{
                $j(i)\leftarrow i$
                \tcp*[r]{ExaPair path}
            }{
                Sample $j(i)$ uniformly from $\mathcal{C}(i)$
                \tcp*[r]{SemPair path}
            }
        }
        Form multimodal pairs $\{(\mathbf{x}_{i}^{a},\mathbf{x}_{j(i)}^{b},y_i):i\in\mathcal{B}\}$\;
        Compute $\mathcal{L}_{\mathrm{cls}}$ and update $\theta,\phi$\;
    }
}

\Return{$f_{\theta^{*}}$}
\end{algorithm*}

The class index requires $O(N)$ preprocessing, while partner lookup introduces negligible overhead relative to network optimization. Pair construction is used only during pretraining, and downstream inference uses the naturally corresponding modalities of each target sample.

For the class-supervised pretraining considered in this work, the original modality-specific and fused classification terms are retained without modification. Denoting them by $\mathcal{L}_{a}$, $\mathcal{L}_{b}$, and $\mathcal{L}_{f}$, respectively, the pretraining objective is
\begin{equation}
\mathcal{L}_{\mathrm{cls}}
=\mathcal{L}_{f}+\lambda_{\mathrm{aux}}(\mathcal{L}_{a}+\mathcal{L}_{b}).
\label{eq:pretraining_loss}
\end{equation}

The loss weighting emphasizes the fused prediction as the primary supervision, while the two modality-specific branches are assigned equal lower weights as auxiliary terms. Because $y_j=y_i$ under SemPair, the supervision remains valid for both modality-specific branches and the fused prediction.

More generally, correspondence relaxation is compatible with objectives whose supervision remains valid after within-class instance replacement. At the population level, MixPair can be represented by the mixture distribution
\begin{equation}
\mathbb{P}_{\rho}
=\rho\mathbb{P}_{\mathrm{E}}
+(1-\rho)\mathbb{P}_{\mathrm{S}}.
\label{eq:mix_distribution}
\end{equation}
This formulation makes $\rho$ a direct control over the relative exposure to ExaPair and SemPair during pretraining.

All partner indices are drawn strictly from the training partition, and validation samples never enter a semantic candidate set. The number of anchors per epoch, data augmentations, input normalization, spatial resolution, optimizer steps, and checkpoint criterion are kept fixed across pairing strategies. SemPair enlarges the set of possible cross-modal associations but does not expose the optimizer to additional anchors or a larger training budget. These controls isolate correspondence granularity from changes in data volume, augmentation strength, or model capacity.

\subsection{Theoretical Analysis of Correspondence Granularity}\label{sec:theory}
Cross-modal correspondence also acts as an inductive bias because it changes which shared factors are repeatedly presented to a multimodal learner. We therefore develop a simple latent-factor model to characterize how exact and relaxed correspondence modify the cross-modal statistical structure observed during pretraining. The analysis is intended as a structural interpretation rather than an exact description of a nonlinear neural network.

\subsubsection{Semantic and Instance Factors}
Consider modalities $a$ and $b$. For instance $i$ with semantic label $y_i$, we decompose each observation into a class-dependent semantic factor, an instance-dependent factor, and a modality-specific residual:
\begin{equation}
\begin{aligned}
\mathbf{x}_{i}^{a}
&=\mathbf{A}_{a}\mathbf{s}_{y_i}
+\mathbf{B}_{a}\mathbf{u}_{i}
+\boldsymbol{\epsilon}_{i}^{a},\\
\mathbf{x}_{i}^{b}
&=\mathbf{A}_{b}\mathbf{s}_{y_i}
+\mathbf{B}_{b}\mathbf{u}_{i}
+\boldsymbol{\epsilon}_{i}^{b}.
\end{aligned}
\label{eq:latent_model}
\end{equation}
Here, $\mathbf{s}_{y_i}$ denotes semantic structure shared by samples of class $y_i$, $\mathbf{u}_{i}$ represents instance-specific properties, and $\boldsymbol{\epsilon}_{i}^{a}$ and $\boldsymbol{\epsilon}_{i}^{b}$ capture modality-specific residual variations. The matrices $\mathbf{A}_{a}$, $\mathbf{A}_{b}$, $\mathbf{B}_{a}$, and $\mathbf{B}_{b}$ describe how the two modalities observe the corresponding latent factors. This additive model is deliberately minimal and does not imply that the nonlinear encoder itself performs such a decomposition. It provides a tractable description of the second-order cross-modal structure induced by different pairing distributions.

We assume that all variables are globally centered and that
\begin{equation}
\mathbb{E}[\mathbf{u}_{i}\mid y_i]=\mathbf{0}.
\label{eq:zero_mean}
\end{equation}
Distinct instances are assumed conditionally independent given their semantic class. The modality-specific residuals are assumed zero-mean and uncorrelated with the semantic and instance factors. Their cross-modal covariance is also assumed zero. These assumptions define $\mathbf{s}_{y}$ as the class-consistent component and place residual within-class variability in $\mathbf{u}_{i}$. Systematic class-related structure is therefore absorbed into $\mathbf{s}_{y}$, whereas pose, local geometry, texture realization, object configuration, or acquisition-specific details may remain in $\mathbf{u}_{i}$.

Under ExaPair, the two observations share both the semantic factor $\mathbf{s}_{y_i}$ and the instance factor $\mathbf{u}_{i}$. Under SemPair, $(\mathbf{x}_{i}^{a},\mathbf{x}_{j}^{b})$ satisfies $i\neq j$ and $y_i=y_j$, so the semantic factor remains common while the instance factors differ.

\subsubsection{SemPair Marginalizes Instance Correlation}
Define the semantic and within-class instance covariances as
\begin{equation}
\boldsymbol{\Sigma}_{s}
=\mathbb{E}[\mathbf{s}_{y}\mathbf{s}_{y}^{\top}],
\qquad
\boldsymbol{\Sigma}_{u}
=\mathbb{E}_{y}\left[\mathbb{E}[\mathbf{u}_{i}\mathbf{u}_{i}^{\top}\mid y]\right].
\label{eq:latent_cov}
\end{equation}

\emph{Proposition 1}: Cross-modal covariance decomposition. Under Eq.~\eqref{eq:latent_model} and the assumptions above, ExaPair induces
\begin{equation}
\boldsymbol{\Sigma}_{ab}^{\mathrm{E}}
=
\underbrace{\mathbf{A}_{a}\boldsymbol{\Sigma}_{s}\mathbf{A}_{b}^{\top}}_{\mathbf{S}}
+
\underbrace{\mathbf{B}_{a}\boldsymbol{\Sigma}_{u}\mathbf{B}_{b}^{\top}}_{\mathbf{U}},
\label{eq:exact_cov}
\end{equation}
whereas SemPair yields
\begin{equation}
\boldsymbol{\Sigma}_{ab}^{\mathrm{S}}=\mathbf{S}.
\label{eq:semantic_cov}
\end{equation}

\textit{Proof}: Under ExaPair, the two modalities contain the same $\mathbf{s}_{y_i}$ and $\mathbf{u}_{i}$. Expanding $\mathbb{E}[\mathbf{x}_{i}^{a}(\mathbf{x}_{i}^{b})^{\top}]$ therefore produces the semantic term $\mathbf{S}$ and the shared instance term $\mathbf{U}$, while the remaining cross terms vanish under the stated assumptions.

Under SemPair, distinct instance factors $\mathbf{u}_{i}$ and $\mathbf{u}_{j}$ are conditionally independent given $y$. Hence,
\begin{equation}
\mathbb{E}[\mathbf{u}_{i}\mathbf{u}_{j}^{\top}\mid y]
=
\mathbb{E}[\mathbf{u}_{i}\mid y]
\mathbb{E}[\mathbf{u}_{j}\mid y]^{\top}
=\mathbf{0},
\end{equation}
which removes the shared instance term while leaving the class-dependent semantic covariance unchanged.

Proposition 1 captures the central statistical distinction between the two pairing strategies. ExaPair repeatedly presents both class-consistent semantic structure and within-instance cross-modal correlation. SemPair preserves the former while marginalizing, in expectation, the correlation supported only by exact instance identity. This statement concerns the pairing distribution presented during pretraining, and it does not imply that a trained nonlinear representation contains no instance-specific information.

The same distinction can be expressed from an information-theoretic perspective under a stronger conditional-independence version of the latent model. Under SemPair, the selected instances and their modality-specific residuals are conditionally independent given the semantic class, and this gives
\begin{equation}
p_{\mathrm{S}}(\mathbf{X}^{a},\mathbf{X}^{b}\mid Y)=p_{\mathrm{S}}(\mathbf{X}^{a}\mid Y)p_{\mathrm{S}}(\mathbf{X}^{b}\mid Y),
\label{eq:conditional_independence}
\end{equation}
and consequently
\begin{equation}
I_{\mathrm{S}}(\mathbf{X}^{a};\mathbf{X}^{b}\mid Y)=0.
\label{eq:conditional_mi}
\end{equation}
In contrast, ExaPair shares the same instance variable $\mathbf{u}_{i}$ across modalities. When this shared instance factor induces non-degenerate dependence in both modalities, the conditional mutual information is positive,
\begin{equation}
I_{\mathrm{E}}(\mathbf{X}^{a};\mathbf{X}^{b}\mid Y)>0.\label{eq:exact_mi}
\end{equation}
SemPair therefore removes within-class cross-modal instance dependence under the assumed model while preserving the shared semantic relation through $Y$.

\subsubsection{A Continuum of Correspondence Granularity}
MixPair induces the population correspondence distribution in Eq.~\eqref{eq:mix_distribution}. Combining this mixture with Proposition 1 gives the following result.

\emph{Corollary 1.}
The cross-modal covariance under instance-pairing ratio $\rho$ is
\begin{equation}
\boldsymbol{\Sigma}_{ab}(\rho)=\mathbf{S}+\rho\mathbf{U}.
\label{eq:mix_cov}
\end{equation}

Equation~\eqref{eq:mix_cov} gives $\rho$ a direct structural interpretation: it controls the contribution of shared instance-specific correlation presented during pretraining. The two endpoints satisfy
\begin{equation}
\rho=0\Longleftrightarrow\mathrm{Semantic\ Pairing},
\qquad
\rho=1\Longleftrightarrow\mathrm{Exact\ Pairing}.
\end{equation}

Intermediate values retain corresponding fractions of the exact-pair component in expectation. The central question is therefore not simply whether cross-modal correspondence is present, but how much exact instance correspondence should be retained.

The epoch-wise global assignment described in Section~\ref{sec:pair_construction} provides a finite-sample realization of this population mixture. Averaging over the random assignments recovers the covariance form in Eq.~\eqref{eq:mix_cov}.

\subsubsection{Transferability and Optimal Correspondence Granularity}
To relate correspondence granularity to downstream transfer, let $\mathbf{M}_{t}$ represent the cross-modal structure relevant to target task $t$, expressed in the same surrogate comparison space as $\mathbf{S}$ and $\mathbf{U}$. We define the structural mismatch as
\begin{equation}
\mathcal{D}_{t}(\rho)
=\left\|\mathbf{S}+\rho\mathbf{U}-\mathbf{M}_{t}\right\|_{F}^{2}.
\label{eq:transfer_discrepancy}
\end{equation}
Equation~\eqref{eq:transfer_discrepancy} is not intended to equal the downstream risk of a nonlinear model. Instead, it provides a surrogate measure of compatibility between the cross-modal structure emphasized during pretraining and the structure relevant to the target task.

\emph{Proposition 2}: Optimal correspondence level. For $\mathbf{U}\neq\mathbf{0}$, the correspondence level minimizing Eq.~\eqref{eq:transfer_discrepancy} is
\begin{equation}
\rho^{*}
=
\Pi_{[0,1]}
\left(
\frac{\left\langle\mathbf{U},\mathbf{M}_{t}-\mathbf{S}\right\rangle_{F}}
{\|\mathbf{U}\|_{F}^{2}}
\right),
\label{eq:optimal_rho}
\end{equation}
where $\Pi_{[0,1]}$ denotes projection onto $[0,1]$. If $\mathbf{U}=\mathbf{0}$, the structural mismatch is independent of $\rho$.

\emph{Proof}: Expanding Eq.~\eqref{eq:transfer_discrepancy} with respect to $\rho$ yields a convex quadratic. Setting its derivative to zero gives the unconstrained optimum
\begin{equation}
\frac{\left\langle\mathbf{U},\mathbf{M}_{t}-\mathbf{S}\right\rangle_{F}}
{\|\mathbf{U}\|_{F}^{2}},
\end{equation}
which is then projected onto the feasible interval $[0,1]$.
Proposition 2 indicates that the preferred correspondence level depends on how strongly the instance-specific structure presented during pretraining matches the structure relevant to the target task. SemPair is favored when this component contributes little to the target-relevant structure, whereas ExaPair is favored when it remains strongly relevant. Intermediate correspondence can be preferable when only part of the instance-specific structure remains relevant. An intuitive special case is
\begin{equation}
\mathbf{M}_{t}=\mathbf{S}+\gamma\mathbf{U},
\qquad
0\leq\gamma\leq1,
\label{eq:target_special}
\end{equation}
for which Eq.~\eqref{eq:optimal_rho} reduces to
\begin{equation}
\rho^{*}=\gamma.
\end{equation}

The preferred correspondence level therefore tracks how much of the instance-specific cross-modal structure remains relevant to the target task. Within this surrogate analysis, increasing the instance-pairing ratio is not guaranteed to reduce the structural mismatch monotonically. This motivates the empirical study in the next section of how downstream transfer changes across different levels of exact instance correspondence.

\section{Experiments}\label{sec:exper}
The proposed pairing-relaxed pretraining framework is evaluated by examining how cross-modal correspondence granularity affects downstream transfer. The experiments first establish overall transferability on four multimodal semantic-segmentation benchmarks, then analyze continuous correspondence granularity and representation structure, and finally test the consistency of the observations across pretraining domains, supervision regimes, adaptation strengths, and model scales. These experiments assess whether relaxed cross-modal correspondence can provide consistently transferable multimodal representations under diverse pretraining and downstream conditions.

\begin{figure}[!t]
\centering
\includegraphics[width=\linewidth]{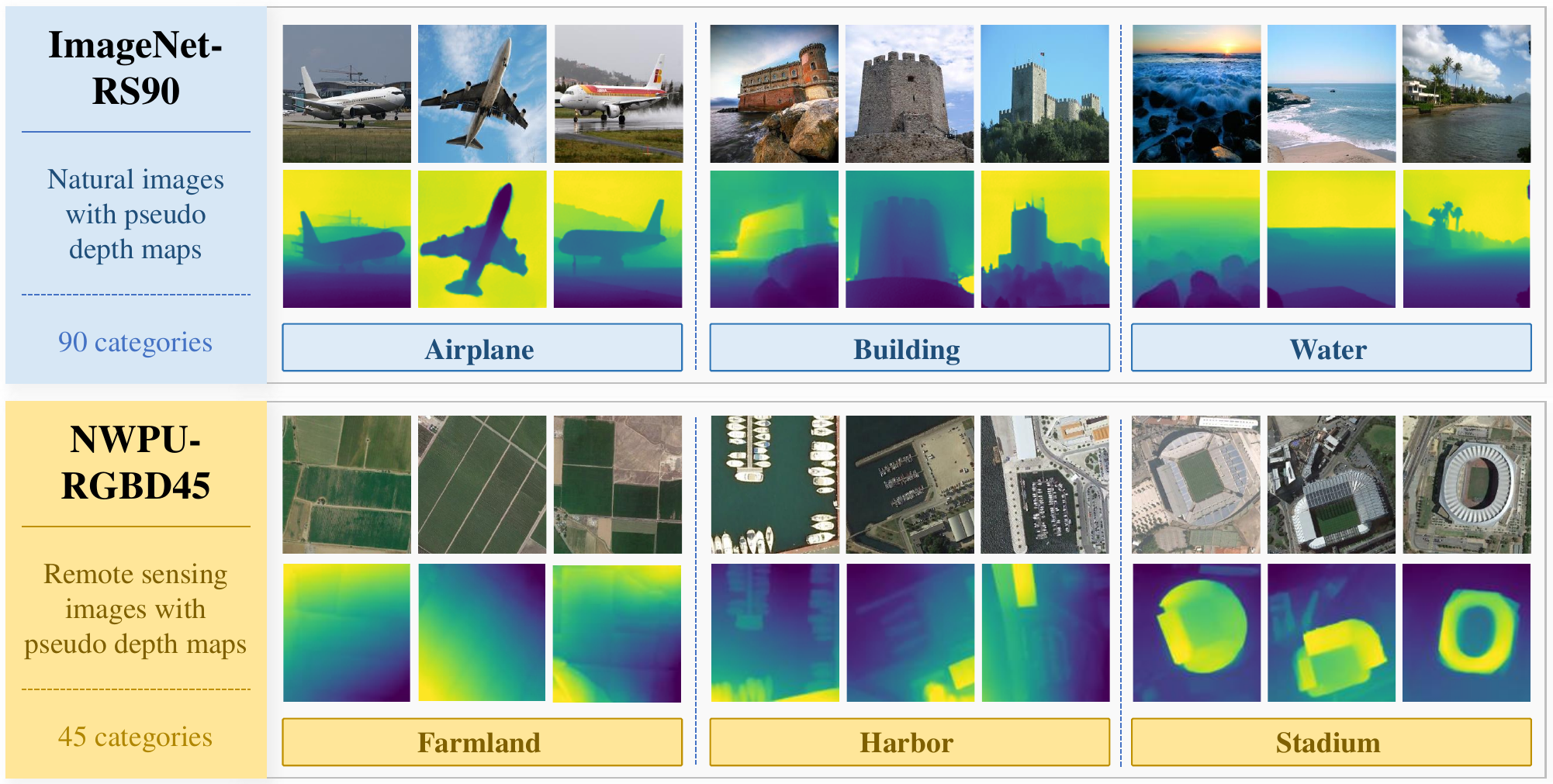}
\caption{Representative samples from the two pretraining domains. ImageNet-RS90 provides an object-centric natural-image source, whereas NWPU-RGBD-45 provides overhead remote-sensing scenes. Both are organized as RGB-pseudo-depth inputs for controlled correspondence experiments.}
\label{fig:pretrain_datasets}
\end{figure}

\begin{table*}[!t]\centering
\caption{Summary of the pretraining datasets. Pseudo-depth maps are generated from RGB images using the same monocular depth estimation pipeline.}\label{tab:pretraining_datasets}
\begin{tabular}{lcccccc}
\toprule
\textbf{Dataset} & \textbf{Source domain} & \textbf{Classes} & \textbf{Pairs} & \textbf{Train} & \textbf{Val.} & \textbf{Input} \\
\midrule
ImageNet-RS90 & ImageNet subset & 90 & 116300 & 110483 & 5817 & $224{\times}224$ \\
NWPU-RGBD-45 & Remote-sensing scenes & 45 & 31500 & 29925 & 1575 & $224{\times}224$ \\
\bottomrule
\end{tabular}
\end{table*}

\subsection{Experimental Setup}\label{sec:exp_setup}
\subsubsection{Pretraining Datasets}
We use two source domains with markedly different image statistics as summarized in Table~\ref{tab:pretraining_datasets}. \textbf{ImageNet-RS90} is a 90-class subset of ImageNet \cite{deng2009imagenet} selected to provide semantically meaningful object-centric categories. After associating each RGB image with a precomputed pseudo-depth map, the corpus contains 116300 samples, of which 110483 are used for training and 5817 for validation. The number of samples per class ranges from 1152 to 1300. This source is intentionally different from overhead remote sensing and therefore provides a useful test of transfer across both task and image domain.

\textbf{NWPU-RGBD-45} is constructed from NWPU-RESISC45 \cite{cheng2017nwpu}. It contains 45 overhead scene categories with 700 images per category, giving 31500 RGB-pseudo-depth samples. We use 29925 for training and 1575 for validation under the same class-wise validation ratio. Compared with ImageNet-RS90, this source is substantially closer to the downstream remote-sensing imagery in viewpoint and scene composition.

For both sources, pseudo-depth is generated once before pretraining by Depth Anything V2 \cite{yang2024depth} and then treated as the second modality. Figure~\ref{fig:pretrain_datasets} presents representative samples from the two pretraining datasets. ExaPair uses the RGB image and the pseudo-depth map of the same instance. SemPair keeps the RGB anchor but draws the pseudo-depth map from another training sample of the same class. MixPair uses the epoch-wise global ratio defined in Section~\ref{sec:pair_construction}, with $\rho=0.5$ by default. Semantic re-pairing is performed only within the training split. RGB-only retains the structure-matched dual-stream backbone while providing RGB information to both streams, following the corresponding unimodal setting of the backbone.

\begin{table*}[!t]\centering
\caption{Summary of the downstream multimodal segmentation benchmarks. GeoNRW uses the Gelsenkirchen subset.}\label{tab:downstream_datasets}\small
\begin{tabular}{lccccccc}
\toprule
\textbf{Dataset} & \textbf{Resolution (m)} & \textbf{Classes} & \textbf{Train} & \textbf{Validation} & \textbf{Test} & \textbf{Tile size} & \textbf{Crop size}\\
\midrule
Potsdam&  0.05& 6&18 & 6 & 14 & $6000\times6000$ & $512\times512$ \\
Vaihingen& 0.09& 6&12 & 4 & 17 & about $2500\times2000$ & $512\times512$ \\
MMHunan& 10& 7&300 & 50 & 150 & $256\times256$ & $256\times256$ \\
GeoNRW& 1& 6&82 & 13 & 42 &  $1000\times1000$ & $512\times512$ \\
\bottomrule
\end{tabular}
\end{table*}
\subsubsection{Downstream Benchmarks}
The pretrained encoders are transferred to four multimodal semantic-segmentation benchmarks of equal status: \textbf{Potsdam}, \textbf{Vaihingen}, \textbf{MMHunan} \cite{li2022dkdfn}, and \textbf{GeoNRW}\cite{GeoNRW}. Table~\ref{tab:downstream_datasets} summarizes the key information of the four downstream benchmarks. They cover different geographic regions, spatial resolutions, scene compositions, and land-cover taxonomies. Optical imagery and DSM are used as the two downstream modalities. Since monocular pseudo-depth and elevation DSM have opposite geometric conventions in our preprocessing pipeline, the DSM is inverted before entering the depth stream to reduce this orientation mismatch. This operation does not make pseudo-depth and DSM physically equivalent.

\begin{figure}[!t]
\centering
\includegraphics[width=\linewidth]{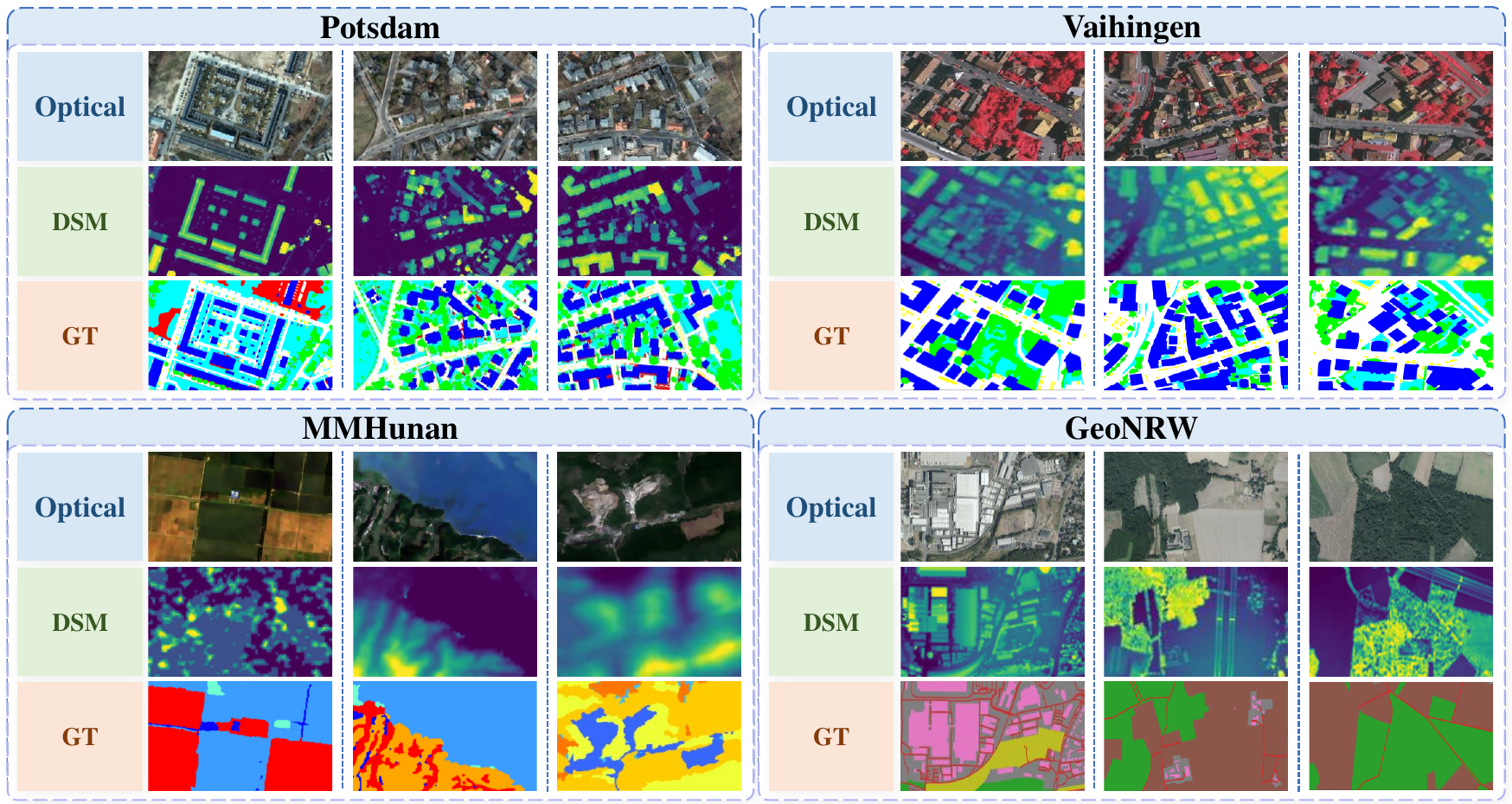}
\caption{Representative multimodal samples and semantic annotations from the downstream benchmarks: (a) Potsdam, (b) Vaihingen, (c) MMHunan, and (d) GeoNRW. Each example shows the optical image, the corresponding elevation representation, and the ground-truth semantic mask.}
\label{fig:downstream_examples}
\end{figure}

Potsdam and Vaihingen additionally support controlled low-label and adaptation analyses. For these two datasets, clutter is excluded from the reported mean IoU and the five foreground categories are used for class-wise comparison. Representative samples from all four datasets are shown in Figure~\ref{fig:downstream_examples}.

\subsubsection{Implementation Details}
\textbf{Pretraining:} We adopt DFormer-Small as the default multimodal encoder unless otherwise stated \cite{yin2026dformer}. Its explicit cross-modal interaction provides a representative backbone for studying correspondence granularity. Pretraining uses $224\times224$ inputs for 50 epochs with AdamW, a learning rate of $10^{-3}$, and weight decay of $0.05$. The learning rate follows a cosine schedule with a five-epoch linear warm-up and a minimum value of $10^{-6}$. The default configuration uses two GPUs with 64 samples per GPU, giving a global batch size of 128. We apply Mixup and CutMix with parameters $0.8$ and $1.0$, respectively, label smoothing of $0.1$, and stochastic depth with a drop-path rate of $0.1$. ExaPair, MixPair, and SemPair use identical settings across all pretraining configurations, with only the cross-modal correspondence strategy varying.

\textbf{Downstream transfer:} The pretraining classification heads are discarded, and the encoder is coupled with the same lightweight segmentation decoder for all strategies. Downstream training uses a batch size of 4 for 50 epochs. We use SGD with momentum $0.9$ and weight decay $5\times10^{-4}$. Standard fine-tuning uses encoder and decoder learning rates of $10^{-3}$ and $10^{-2}$, respectively. Both learning rates are reduced by a factor of $0.1$ at epochs 25, 35, and 45. Frozen transfer fixes all encoder parameters and trains only the decoder, whereas the scratch baseline trains both components from random initialization with a learning rate of $10^{-2}$. The downstream protocol is held fixed across pairing strategies within each benchmark and adaptation setting.

\subsubsection{Evaluation Protocol}
For semantic segmentation, we report overall accuracy (OA), mean F1 score (mF1), Cohen's kappa, class-wise intersection over union (IoU), and mean IoU (mIoU). Each configuration is independently run \textbf{twice} with the same protocol, and the better of the two runs is reported uniformly for all methods. All quantitative downstream results are reported on the validation split. Within each run, the checkpoint with the highest validation mIoU is selected, and all accompanying metrics are computed on the same validation split using that checkpoint.

For representation-level retrieval, we use the fixed validation split of each pretraining dataset and keep the original full DFormer forward path unchanged. RGB-stream features are used as queries and depth-stream features as the gallery. Instance R@1 uses the complete gallery and asks whether the exact depth counterpart is ranked first. For semantic retrieval, the exact same-instance depth sample is removed from the gallery before computing semantic R@1 and R@5. These metrics therefore measure whether a different instance of the same class can be retrieved. As DFormer is a jointly interactive encoder, these results are interpreted as properties of the complete learned multimodal representation rather than as measurements of isolated unimodal encoders.

\begin{table*}[!t]\centering
\caption{Pretraining validation results. All metrics in a row are taken from the epoch with the highest fused-branch Acc@1.}\label{tab:pretrain_results}
\begin{tabular}{llccccc}
\toprule
\textbf{Dataset} & \textbf{Pretraining} & \textbf{Val. loss $\downarrow$} & \textbf{Acc@1 $\uparrow$} & \textbf{Acc@5 $\uparrow$} & \textbf{RGB Acc@1 $\uparrow$} & \textbf{Depth Acc@1 $\uparrow$} \\
\midrule
\multirow{4}{*}{ImageNet-RS90}
& RGB-only & \textbf{0.8060} & \textbf{82.50} & \textbf{97.04} & \textbf{82.12} & / \\
& ExaPair   & 0.8398 & 81.80 & 96.80 & 81.78 & \textbf{80.85} \\
& MixPair  & 0.8509 & 81.08 & 96.42 & 81.20 & 79.84 \\
& SemPair  & 0.9086 & 79.25 & 95.65 & 79.13 & 78.53 \\
\midrule
\multirow{4}{*}{NWPU-RGBD-45}
& RGB-only & 0.5563 & 90.04 & 99.18 & 89.85 & / \\
& ExaPair   & \textbf{0.4783} & \textbf{91.88} & 99.18 & \textbf{91.62} & \textbf{91.37} \\
& MixPair  & 0.5060 & 90.80 & \textbf{99.30} & 90.93 & 90.29 \\
& SemPair  & 0.5676 & 88.32 & 98.98 & 87.94 & 88.07 \\
\bottomrule
\end{tabular}
\end{table*}

\begin{table*}[!t]\centering
\caption{Primary transfer results on Potsdam. Imp., Build., and Low veg. denote impervious surfaces, buildings, and low vegetation, respectively.}\label{tab:primary_potsdam}
\begin{tabular}{llccccccccc}
\toprule
\multirow{2}{*}{\textbf{Adaptation}} & \multirow{2}{*}{\textbf{Pretraining}} & \multirow{2}{*}{\textbf{OA}} & \multirow{2}{*}{\textbf{mF1}} & \multirow{2}{*}{\textbf{Kappa}} & \multicolumn{5}{c}{\textbf{Class-wise IoU}} & \multirow{2}{*}{\textbf{mIoU}} \\
\cmidrule(lr){6-10}
 & & & & & \textbf{Imp.} & \textbf{Build.} & \textbf{Low veg.} & \textbf{Tree} & \textbf{Car} & \\
\midrule
Scratch & None & 78.74 & 76.83 & 71.79 & 72.78 & 87.19 & 54.53 & 43.58 & 60.61 & 63.74 \\
\midrule
\multirow{4}{*}{Frozen}
& RGB-only & 84.36 & 85.39 & 79.31 & 76.41 & 90.31 & 63.62 & 64.55 & 80.49 & 75.07 \\
& ExaPair  & 85.68 & 86.58 & 81.15 & 79.41 & \textbf{91.08} & 65.21 & 65.41 & 83.53 & 76.93 \\
& MixPair & \textbf{85.81} & \textbf{86.77} & \textbf{81.27} & 79.58 & 90.94 & \textbf{65.68} & 65.63 & \textbf{84.26} & \textbf{77.22} \\
& SemPair & 85.72 & 86.67 & 81.16 & \textbf{79.72} & 91.03 & 64.69 & \textbf{66.06} & 83.79 & 77.06 \\
\midrule
\multirow{4}{*}{Fine-tuning}
& RGB-only & 87.90 & 88.84 & 84.08 & 82.79 & 91.94 & 69.25 & 70.89 & 86.98 & 80.37 \\
& ExaPair  & 88.04 & 88.97 & 84.28 & 82.89 & \textbf{91.98} & 70.04 & 70.49 & \textbf{87.42} & 80.57 \\
& MixPair & \textbf{88.26} & \textbf{89.05} & \textbf{84.53} & \textbf{83.28} & 91.89 & 70.12 & 70.94 & 87.16 & \textbf{80.68} \\
& SemPair & 88.18 & 89.03 & 84.44 & 83.13 & 91.88 & \textbf{70.18} & \textbf{71.11} & 86.89 & 80.64 \\
\bottomrule
\end{tabular}
\end{table*}

\subsection{Overall Transferability on Multimodal Benchmarks}\label{sec:overall_transfer}
We first test the central claim directly: does a multimodal encoder remain transferable when exact instance pairing is removed during pretraining? Table~\ref{tab:pretrain_results} reports source validation performance during pretraining, and Tables~\ref{tab:primary_potsdam}-\ref{tab:primary_geonrw} report downstream segmentation under frozen transfer and standard fine-tuning.

\subsubsection{Source Fitting Versus Downstream Transfer}
The pretraining validation results in Table~\ref{tab:pretrain_results} show a consistent advantage of ExaPair over MixPair and SemPair in source-domain classification. The results on ImageNet-RS90 indicate that ExaPair achieves a fused Acc@1 of $81.80\%$, compared with $81.08\%$ for MixPair and $79.25\%$ for SemPair. A similar trend is observed on NWPU-RGBD-45, where the corresponding accuracies are $91.88\%$, $90.80\%$, and $88.32\%$. These results indicate that preserving exact instance correspondence facilitates source-domain fitting, as expected from the additional cross-modal instance information available to ExaPair. Nevertheless, relaxing correspondence does not lead to optimization failure or representation collapse: both MixPair and SemPair retain strong classification performance on the two pretraining datasets. This suggests that exact instance correspondence is beneficial, but not indispensable, for learning meaningful multimodal representations during pretraining.

\subsubsection{Primary Results on Four Multimodal Benchmarks}
Tables~\ref{tab:primary_potsdam}-\ref{tab:primary_geonrw} report the primary transfer results on four multimodal semantic-segmentation benchmarks under both frozen-encoder evaluation and standard end-to-end fine-tuning. The frozen setting provides a direct assessment of the pretrained representation because the downstream decoder must operate on a fixed encoder, whereas fine-tuning allows the pretrained representation to adapt to the target task.

Before comparing the three cross-modal pairing strategies, the baseline results provide two useful sanity checks. First, RGB-only pretraining consistently improves over training from scratch, confirming the effectiveness of the pretraining-and-transfer protocol itself. Second, the three multimodal strategies generally provide further gains over the RGB-only initialization, showing that the pseudo-depth modality contributes useful complementary information rather than merely increasing the input complexity. These observations establish a meaningful multimodal transfer setting in which the effect of cross-modal correspondence can be examined independently.

\textbf{Potsdam:} As shown in Table~\ref{tab:primary_potsdam}, ExaPair, MixPair, and SemPair achieve $76.93\%$, $77.22\%$, and $77.06\%$ mIoU, respectively, compared with $75.07\%$ for RGB-only pretraining. Notably, removing exact instance correspondence in SemPair does not reduce frozen transfer performance. SemPair's mIoU is slightly higher than that of ExaPair, and MixPair gives the best overall result. After fine-tuning, the three pairing strategies become even closer, reaching $80.57\%$, $80.68\%$, and $80.64\%$ mIoU. Their spread is only $0.11\%$, indicating that the transferability of the pretrained representation is largely preserved across substantially different correspondence levels. The class-wise results show a similar pattern. SemPair achieves the highest IoUs for low vegetation and trees, while MixPair obtains the highest overall mIoU, suggesting that the effect is not driven by a single semantic category.

\textbf{Vaihingen:} The results on Vaihingen in Table~\ref{tab:primary_vaihingen} provide an even clearer comparison under frozen transfer. RGB-only pretraining reaches $67.14\%$ mIoU, whereas ExaPair, MixPair, and SemPair improve it to $70.16\%$, $71.53\%$, and $70.61\%$, respectively. SemPair therefore exceeds ExaPair by $0.45\%$ despite never observing exact cross-modal instance pairs during pretraining, while MixPair achieves the highest frozen performance. After fine-tuning, the differences remain small: ExaPair obtains $74.13\%$, MixPair $74.52\%$, and SemPair $73.94\%$ mIoU. Although SemPair is $0.19\%$ below ExaPair in this setting, both remain above the RGB-only baseline of $72.49\%$. The class-wise results also show no systematic dominance of ExaPair: ExaPair performs best on impervious surfaces and buildings, MixPair on low vegetation and cars, and SemPair on trees. This distribution of gains suggests that correspondence relaxation does not simply favor or suppress a particular land-cover category.

\begin{table*}[!t]\centering
\caption{Primary transfer results on Vaihingen. Imp., Build., and Low veg. denote impervious surfaces, buildings, and low vegetation, respectively.}\label{tab:primary_vaihingen}
\begin{tabular}{llccccccccc}
\toprule
\multirow{2}{*}{\textbf{Adaptation}} & \multirow{2}{*}{\textbf{Pretraining}} & \multirow{2}{*}{\textbf{OA}} & \multirow{2}{*}{\textbf{mF1}} & \multirow{2}{*}{\textbf{Kappa}} & \multicolumn{5}{c}{\textbf{Class-wise IoU}} & \multirow{2}{*}{\textbf{mIoU}} \\
\cmidrule(lr){6-10}
 & & & & & \textbf{Imp.} & \textbf{Build.} & \textbf{Low veg.} & \textbf{Tree} & \textbf{Car} & \\
\midrule
Scratch & None & 82.67 & 72.19 & 76.69 & 71.81 & 81.12 & 50.02 & 74.51 & 21.71 & 59.83 \\
\midrule
\multirow{4}{*}{Frozen}
& RGB-only & 84.03 & 79.61 & 78.50 & 71.22 & 81.72 & 52.66 & 78.90 & 51.23 & 67.14 \\
& ExaPair  & 85.80 & 81.93 & 80.89 & 74.35 & 82.74 & 58.13 & 80.33 & 55.25 & 70.16 \\
& MixPair & \textbf{86.63} & \textbf{82.88} & \textbf{82.01} & \textbf{75.89} & \textbf{84.67} & 59.00 & \textbf{80.96} & \textbf{57.12} & \textbf{71.53} \\
& SemPair & 86.20 & 82.22 & 81.42 & 74.83 & 83.49 & \textbf{59.40} & 80.80 & 54.51 & 70.61 \\
\midrule
\multirow{4}{*}{Fine-tuning}
& RGB-only & 87.38 & 83.48 & 83.02 & 78.11 & 87.44 & 59.09 & 80.29 & 57.56 & 72.49 \\
& ExaPair  & \textbf{88.83} & 84.52 & \textbf{84.94} & \textbf{81.47} & \textbf{89.60} & 60.56 & 81.38 & 57.66 & 74.13 \\
& MixPair & 88.31 & \textbf{84.95} & 84.28 & 79.71 & 87.88 & \textbf{62.08} & 81.50 & \textbf{61.41} & \textbf{74.52} \\
& SemPair & 88.36 & 84.49 & 84.32 & 79.94 & 88.13 & 61.31 & \textbf{81.57} & 58.74 & 73.94 \\
\bottomrule
\end{tabular}
\end{table*}

\begin{table*}[!t]\centering
\caption{Primary transfer results on MMHunan. Crop., Grass., Wet., and Unused denote cropland, grassland, wetland, and unused land, respectively.}\label{tab:primary_mmhunan}
\begin{tabular}{llccccccccccc}
\toprule
\multirow{2}{*}{\textbf{Adaptation}} & \multirow{2}{*}{\textbf{Pretraining}} & \multirow{2}{*}{\textbf{OA}} & \multirow{2}{*}{\textbf{mF1}} & \multirow{2}{*}{\textbf{Kappa}} & \multicolumn{7}{c}{\textbf{Class-wise IoU}} & \multirow{2}{*}{\textbf{mIoU}} \\
\cmidrule(lr){6-12}
 & & & & & \textbf{Crop.} & \textbf{Forest} & \textbf{Grass.} & \textbf{Wet.} & \textbf{Water} & \textbf{Unused} & \textbf{Built-up} & \\
\midrule
Scratch & None & 68.04 & 43.90 & 55.01 & 45.69 & 68.35 & 10.68 & 0.00 & 60.70 & 22.33 & 19.07 & 32.40 \\
\midrule
\multirow{4}{*}{Frozen}
& RGB-only & 72.96 & 50.72 & 61.72 & 47.79 & 71.91 & 13.24 & 0.32 & 65.36 & 51.77 & 21.52 & 38.84 \\
& ExaPair & 74.14 & \textbf{53.46} & 63.54 & 51.48 & 71.97 & \textbf{14.06} & \textbf{4.91} & 70.54 & 51.65 & \textbf{23.24} & \textbf{41.12} \\
& MixPair & \textbf{74.64} & 53.13 & \textbf{64.27} & \textbf{52.04} & \textbf{72.47} & 13.33 & 4.62 & \textbf{71.03} & 52.54 & 21.30 & 41.05 \\
& SemPair & 74.36 & 52.95 & 63.80 & 51.46 & 72.19 & 13.38 & 3.60 & 71.00 & \textbf{53.14} & 21.84 & 40.94 \\
\midrule
\multirow{4}{*}{Fine-tuning}
& RGB-only & 74.51 & 53.70 & 64.28 & 51.16 & \textbf{74.01} & 17.11 & 1.91 & 66.80 & 53.28 & \textbf{25.45} & 41.39 \\
& ExaPair & 74.23 & 54.28 & 63.91 & 51.81 & 72.40 & 15.59 & \textbf{5.94} & 69.55 & 53.77 & 23.09 & 41.74 \\
& MixPair & 75.22 & \textbf{55.46} & \textbf{65.27} & 52.85 & 73.04 & \textbf{17.32} & 5.83 & \textbf{73.67} & 53.02 & 24.98 & \textbf{42.96} \\
& SemPair & \textbf{75.24} & 54.69 & 65.22 & \textbf{53.10} & 73.32 & 15.00 & 5.91 & 72.59 & \textbf{53.90} & 23.02 & 42.41 \\
\bottomrule
\end{tabular}
\end{table*}

\begin{table*}[!t]\centering
\caption{Primary transfer results on GeoNRW. Agric. and Grass. denote agricultural areas and grassland, respectively.}\label{tab:primary_geonrw}
\begin{tabular}{llcccccccccc}
\toprule
\multirow{2}{*}{\textbf{Adaptation}} & \multirow{2}{*}{\textbf{Pretraining}} & \multirow{2}{*}{\textbf{OA}} & \multirow{2}{*}{\textbf{mF1}} & \multirow{2}{*}{\textbf{Kappa}} & \multicolumn{6}{c}{\textbf{Class-wise IoU}} & \multirow{2}{*}{\textbf{mIoU}} \\
\cmidrule(lr){6-11}
 & & & & & \textbf{Forest} & \textbf{Agric.} & \textbf{Built-up} & \textbf{Grass.} & \textbf{Road} & \textbf{Building} & \\
\midrule
Scratch & None & 67.39 & 54.01 & 52.15 & 31.07 & 70.26 & 60.89 & 12.21 & 14.92 & 54.72 & 40.68 \\
\midrule
\multirow{4}{*}{Frozen}
& RGB-only & 69.15 & 57.94 & 54.10 & 38.25 & 74.65 & 62.69 & 19.41 & 18.00 & 50.10 & 43.85 \\
& ExaPair & 70.68 & 61.22 & 57.44 & 39.04 & 76.52 & 63.96 & 20.62 & \textbf{26.02} & 55.01 & 46.86 \\
& MixPair & \textbf{71.69} & 61.80 & \textbf{58.76} & 39.86 & \textbf{77.72} & \textbf{65.05} & 20.99 & 25.54 & 56.40 & 47.60 \\
& SemPair & 71.42 & \textbf{62.14} & 58.41 & \textbf{40.51} & 76.72 & 64.49 & \textbf{21.71} & 25.91 & \textbf{57.59} & \textbf{47.82} \\
\midrule
\multirow{4}{*}{Fine-tuning}
& RGB-only & 72.36 & 63.34 & 59.99 & 42.18 & 78.29 & 65.83 & \textbf{26.58} & 24.03 & 57.18 & 49.01 \\
& ExaPair & 73.04 & 63.67 & 60.73 & 41.39 & 78.79 & 66.83 & 22.53 & 28.50 & 58.84 & 49.48 \\
& MixPair & \textbf{73.42} & 63.88 & \textbf{61.27} & 41.68 & \textbf{79.66} & \textbf{67.39} & 23.50 & 26.96 & 59.60 & 49.80 \\
& SemPair & 73.20 & \textbf{64.43} & \textbf{61.27} & \textbf{42.68} & 76.69 & 67.05 & 24.81 & \textbf{28.61} & \textbf{60.51} & \textbf{50.06} \\
\bottomrule
\end{tabular}
\end{table*}

\textbf{MMHunan:} Table~\ref{tab:primary_mmhunan} reports transfer results with mIoU and mF1 averaged over all seven land-cover classes. Under frozen transfer, ExaPair, MixPair, and SemPair obtain $41.12\%$, $41.05\%$, and $40.94\%$ mIoU, respectively, compared with $38.84\%$ for RGB-only. Their spread is only $0.18\%$, indicating that SemPair preserves performance close to ExaPair. After fine-tuning, the corresponding mIoUs are $41.74\%$, $42.96\%$, and $42.41\%$. These differences remain modest and do not establish a universal advantage for any pairing strategy. At the selected fine-tuning epochs, ExaPair has the highest wetland IoU among the three multimodal strategies, MixPair has the highest grassland, water, and built-up IoUs, and SemPair has the highest cropland, forest, and unused-land IoUs.

\textbf{GeoNRW:} On GeoNRW (Table~\ref{tab:primary_geonrw}), SemPair reaches $47.82\%$ mIoU under frozen transfer, compared with $46.86\%$ for ExaPair, $47.60\%$ for MixPair, and $43.85\%$ for RGB-only. After fine-tuning, SemPair obtains $50.06\%$, MixPair $49.80\%$, and ExaPair $49.48\%$. The differences between the multimodal strategies are modest, with SemPair remaining competitive with ExaPair. Among the three multimodal strategies, SemPair has the highest fine-tuned IoUs for forest, grassland, road, and building, while MixPair has the highest agricultural and built-up IoUs.

\begin{figure*}[!t]
\centering
\includegraphics[width=\linewidth]{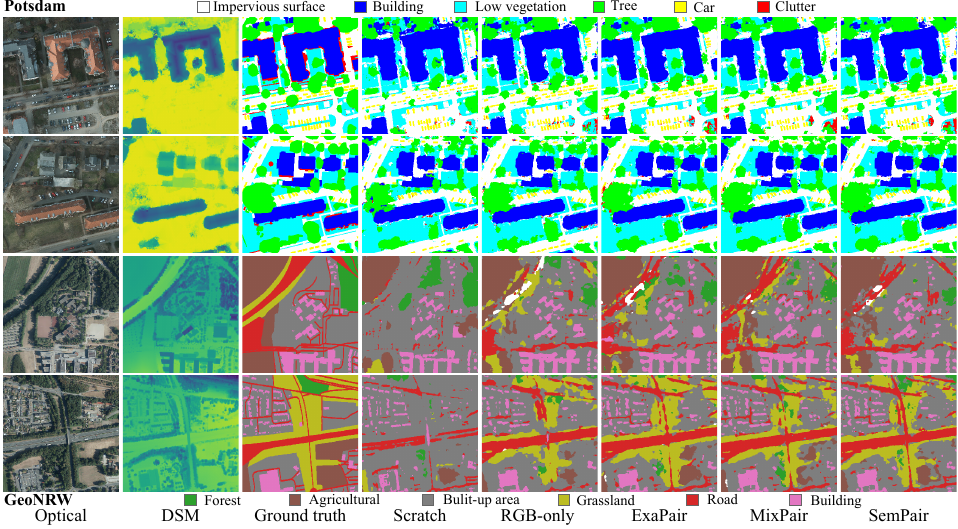}
\caption{Qualitative segmentation results on test set from Potsdam ($2048\times2048$) and GeoNRW ($512\times512$). Each row shows the optical image, DSM, ground truth, and predictions from the end-to-end trained Scratch baseline and four pretrained models under frozen transfer.}
\label{fig:downstream_visualization}
\end{figure*}

Figure~\ref{fig:downstream_visualization} provides qualitative comparisons on Potsdam and GeoNRW. The pretrained models generally recover more coherent semantic regions than Scratch, while ExaPair, MixPair, and SemPair exhibit broadly similar spatial predictions with local differences in boundaries and category assignments. These examples complement the quantitative evidence that relaxing exact instance pairing can preserve downstream segmentation quality.

Overall, the four benchmarks show a consistent pattern. Across the eight frozen and fine-tuned evaluation settings, SemPair remains within $0.20\%$ of ExaPair in the two cases where it is lower, while giving slightly higher mIoU in the remaining six settings. This includes frozen transfer, where encoder adaptation cannot compensate for differences introduced during pretraining. These observations support comparable transfer performance from class-consistent cross-instance pretraining. MixPair is also competitive, indicating that preserving more exact instance correspondence does not systematically improve downstream transfer.

These downstream results also provide an informative contrast to the source-domain classification results in Table~\ref{tab:pretrain_results}, where ExaPair consistently achieves higher Acc@1 than SemPair. The advantage of exact correspondence for source fitting is therefore not preserved as a comparable advantage after transfer. This observation is consistent with the central premise of our analysis: exact instance correspondence can provide additional source-domain cross-modal dependence, but retaining that dependence is not a prerequisite for effective downstream transfer.

\begin{table*}[!t]\centering
\caption{Effect of continuous correspondence granularity on downstream mF1 and mIoU (\%).}\label{tab:pairing_granularity}
\begin{tabular}{llcccccccc}
\toprule
\multirow{3}{*}{\textbf{$\rho$}} & \multirow{3}{*}{\textbf{Strategy}} & \multicolumn{4}{c}{\textbf{Potsdam}} & \multicolumn{4}{c}{\textbf{Vaihingen}} \\
\cmidrule(lr){3-6}\cmidrule(lr){7-10}
 &  & \multicolumn{2}{c}{\textbf{Frozen}} & \multicolumn{2}{c}{\textbf{Fine-tuning}} & \multicolumn{2}{c}{\textbf{Frozen}} & \multicolumn{2}{c}{\textbf{Fine-tuning}} \\
\cmidrule(lr){3-4}\cmidrule(lr){5-6}\cmidrule(lr){7-8}\cmidrule(lr){9-10}
 &  & \textbf{mF1} & \textbf{mIoU} & \textbf{mF1} & \textbf{mIoU} & \textbf{mF1} & \textbf{mIoU} & \textbf{mF1} & \textbf{mIoU} \\
\midrule
0 & SemPair & 86.67 & 77.06 & 89.03 & 80.64 & 82.22 & 70.61 & 84.49 & 73.94 \\
0.25 & MixPair & 86.68 & 77.10 & 88.98 & 80.59 & 82.32 & 70.76 & 84.77 & 74.40 \\
0.50 & MixPair & \textbf{86.77} & \textbf{77.22} & \textbf{89.05} & \textbf{80.68} & \textbf{82.88} & \textbf{71.53} & \textbf{84.95} & \textbf{74.52} \\
0.75 & MixPair & 86.58 & 76.94 & 88.98 & 80.58 & 81.75 & 69.92 & 84.09 & 73.40 \\
1 & ExaPair & 86.58 & 76.93 & 88.97 & 80.57 & 81.93 & 70.16 & 84.52 & 74.13 \\
\bottomrule
\end{tabular}
\end{table*}

\subsection{Understanding Cross-Modal Correspondence Granularity}\label{sec:granularity_analysis}
The overall results establish that ExaPair can be removed without significantly degrading transfer. We next ask a more general question: does transfer improve monotonically as more exact correspondence is restored?

\subsubsection{Continuous Correspondence Granularity}
We vary the epoch-level instance-pairing ratio over $\rho\in\{0,0.25,0.5,0.75,1\}$. Here $\rho=0$ corresponds to SemPair, $\rho=1$ to ExaPair, and $\rho=0.5$ to the default MixPair. At every epoch, $N_{\mathrm{E}}=\lfloor\rho N\rfloor$ anchors are assigned to ExaPair after a random permutation of the training set, while the remaining anchors use SemPair. No class-wise stratification is used, and all other conditions remain fixed.

\begin{figure}[!t]
\centering
{\includegraphics[width=\linewidth]{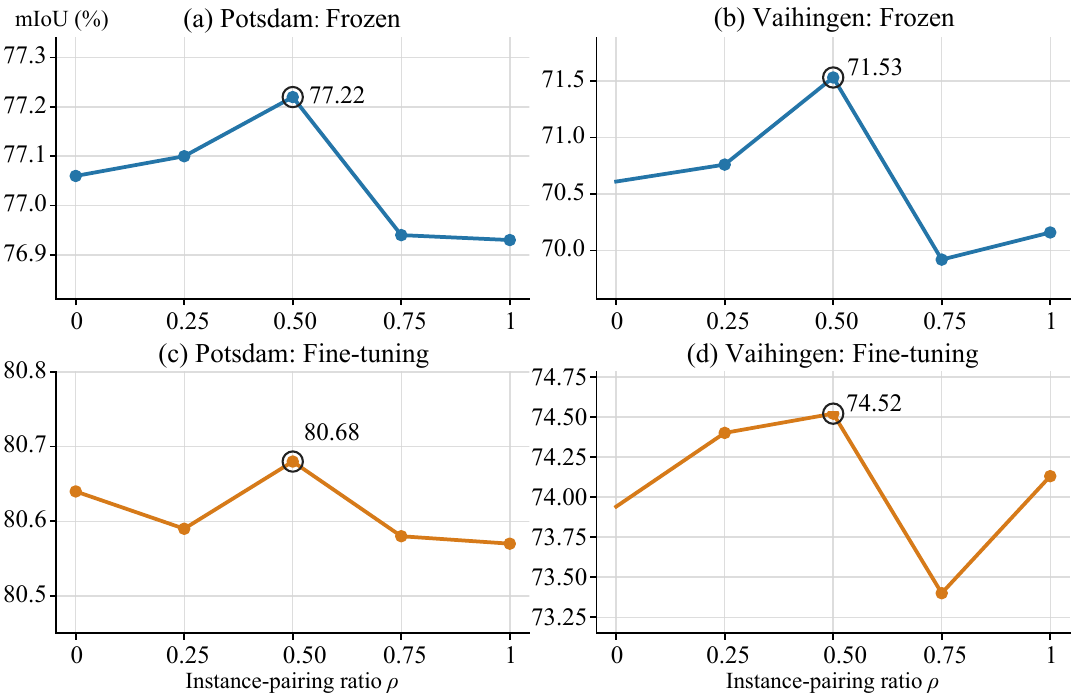}}
\caption{Effect of the instance-pairing ratio $\rho$ on downstream mIoU for Potsdam and Vaihingen under frozen transfer and end-to-end fine-tuning.}
\label{fig_pairing_granularity}
\end{figure}

Table~\ref{tab:pairing_granularity} and Figure~\ref{fig_pairing_granularity} examine how the instance-pairing ratio $\rho$ affects downstream transfer. The Potsdam results exhibit limited sensitivity to correspondence granularity, with mIoU varying by only $0.29\%$ and $0.11\%$ under frozen transfer and fine-tuning, respectively. In contrast, Vaihingen shows more pronounced variations, particularly under frozen transfer, where the mIoU increases from $70.61\%$ at $\rho=0$ to $71.53\%$ at $\rho=0.5$, before decreasing to $69.92\%$ at $\rho=0.75$. A similar non-monotonic trend is observed after fine-tuning. This difference in sensitivity may be partly attributed to the larger downstream training set of Potsdam, which provides more supervision for adapting pretrained representations, whereas the more limited supervision in Vaihingen may make the inherited pretraining characteristics more influential.

Despite these differences, $\rho=0.5$ consistently achieves the highest mIoU among the evaluated ratios across both datasets and adaptation settings. This observation suggests that an equal mixture of ExaPair and SemPair may provide a favorable balance between preserving instance-specific cross-modal information and encouraging shared semantic structure. Such a balance is consistent with the latent-factor analysis in Eq.~\eqref{eq:mix_cov}, where $\rho$ controls the contribution of instance-specific cross-modal correlation. Nevertheless, the overall performance differences among the pairing strategies are generally modest, particularly on Potsdam. Rather than establishing a universally optimal ratio, these results indicate that increasing exact instance correspondence does not necessarily improve downstream transfer, and that moderate correspondence relaxation can maintain competitive performance.

\subsubsection{Representation-Level Cross-Modal Retrieval}\label{sec:representation_evidence}
This analysis examines whether SemPair preserves cross-modal information despite relaxing exact instance correspondence during pretraining. Exact-instance retrieval and instance-excluded semantic retrieval are evaluated between the RGB and depth branches of the pretrained encoder. Features are extracted from paired RGB-depth inputs with cross-modal interaction active, so the evaluation measures cross-branch correspondence under multimodal conditioning rather than retrieval between independently encoded modalities.

As presented in Table~\ref{tab:representation_retrieval}, all three strategies achieve high exact-instance R@1 with the complete pretrained DFormer. SemPair reaches $97.59\%$ on ImageNet-RS90, only $0.30\%$ below ExaPair, and $96.13\%$ on NWPU-RGBD-45, compared with $97.40\%$ for ExaPair. Instance-excluded retrieval provides a complementary assessment of semantic organization by removing the exact counterpart from the candidate gallery. SemPair achieves semantic R@1 and R@5 of $72.80\%$ and $88.95\%$ on ImageNet-RS90, trailing ExaPair by only $1.40\%$ and $1.46\%$, respectively. Comparable gaps of $1.40\%$ and $1.27\%$ are observed on NWPU-RGBD-45, where MixPair also achieves the highest semantic R@1. These results indicate that SemPair retains much of the cross-modal structure measured by retrieval, even without exact instance pairs during pretraining. Together with the downstream segmentation results, they provide complementary evidence for the transferability of representations learned through semantic-level correspondence.

\begin{figure}[!t]
\centering
\includegraphics[width=\linewidth]{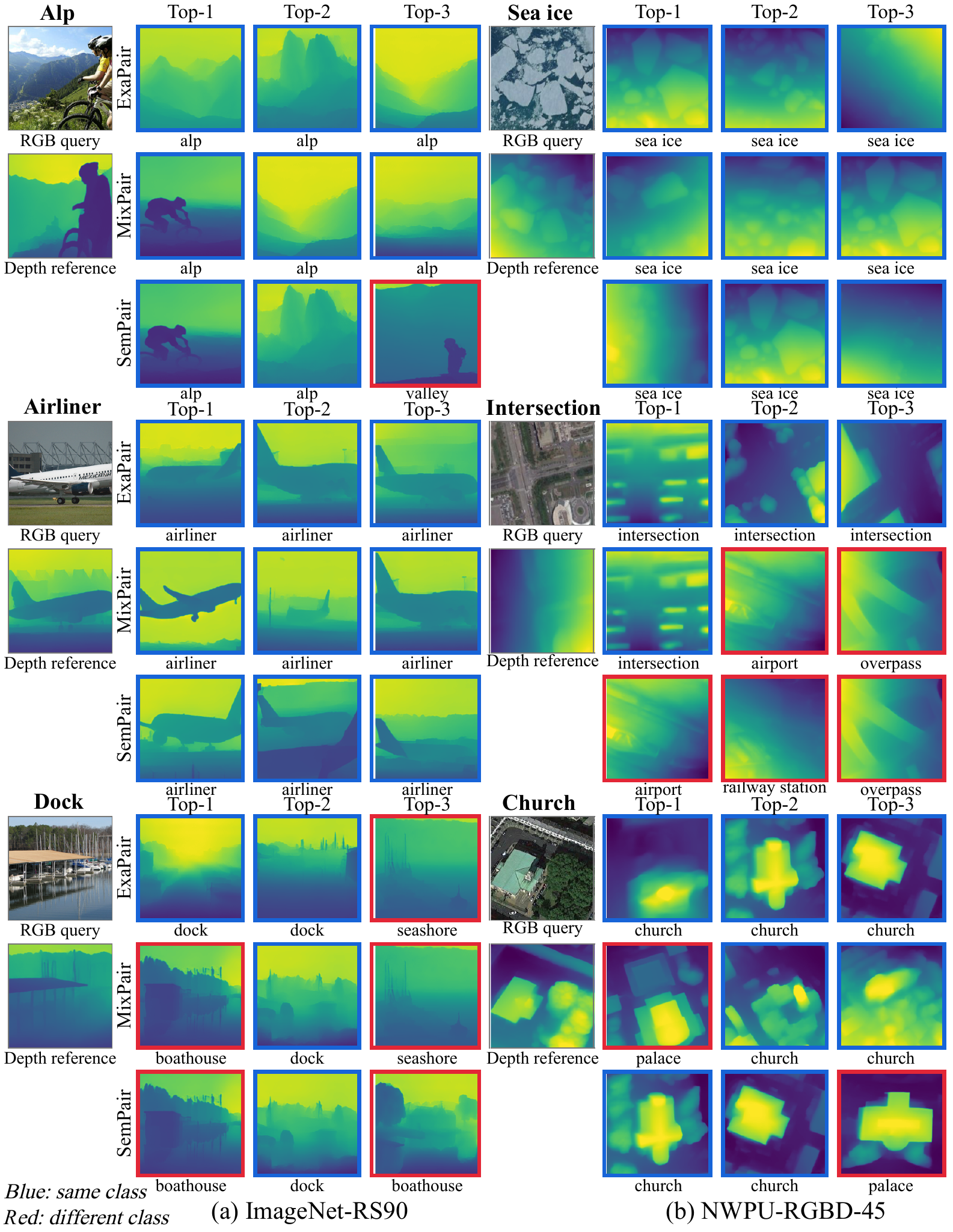}
\caption{Qualitative cross-branch RGB-to-depth retrieval on the fixed validation splits of (a) ImageNet-RS90 and (b) NWPU-RGBD-45. Each example shows the RGB query and its same-instance depth reference once, followed by the top-three retrieved depth samples for ExaPair, MixPair, and SemPair. The reference depth is excluded from the retrieval gallery.}
\label{fig:retrieval_visualization}
\end{figure}

\begin{table}[!t]\centering
\caption{Representation-level RGB-to-depth retrieval on the fixed pretraining validation split. Inst. R@1 uses the full depth gallery. Excl-inst. Sem. R@1 and R@5 exclude the same-instance depth sample from the gallery. All metrics are reported in percentages.}\label{tab:representation_retrieval}
\footnotesize
\setlength{\tabcolsep}{2.5pt}
\begin{tabular}{>{\raggedright\arraybackslash}p{0.24\columnwidth}>{\raggedright\arraybackslash}p{0.16\columnwidth}>{\centering\arraybackslash}p{0.12\columnwidth}>{\centering\arraybackslash}p{0.19\columnwidth}>{\centering\arraybackslash}p{0.19\columnwidth}}
\toprule
\textbf{Pretraining domain}& \textbf{Strategy}& \textbf{Inst. R@1}& \textbf{Excl-inst. Sem. R@1}& \textbf{Excl-inst. Sem. R@5} \\
\midrule
\multirow{3}{=}{ImageNet-RS90}& ExaPair & \textbf{97.89} & \textbf{74.20} & \textbf{90.41} \\
& MixPair & 97.09 & 72.98 & 89.77 \\
& SemPair & 97.59 & 72.80 & 88.95 \\
\midrule
\multirow{3}{=}{NWPU-RGBD-45}& ExaPair & \textbf{97.40} & 86.86 & \textbf{96.06} \\
& MixPair & 96.44 & \textbf{87.75} & 95.81 \\
& SemPair & 96.13 & 85.46 & 94.79 \\
\bottomrule
\end{tabular}
\end{table}
Figure~\ref{fig:retrieval_visualization} presents representative instance-excluded retrieval examples from both pretraining domains. All three strategies retrieve semantically consistent samples for categories such as airliner and sea ice, whereas more challenging categories, including dock and intersection, exhibit differences in retrieval rankings and category confusions. These results further demonstrate that SemPair can learn meaningful cross-modal semantic associations without exact instance correspondence during pretraining.

\subsection{Generalization Across Pretraining Domains}\label{sec:pretraining_domain_generalization}
The preceding representation analysis uses both pretraining sources, whereas the primary transfer experiments use ImageNet-RS90. We therefore repeat downstream transfer after pretraining on NWPU-RGBD-45 to examine whether the observations depend on a particular source domain.

\begin{table*}[!t]\centering
\caption{Generalization across pretraining domains measured by downstream mF1 and mIoU (\%).}\label{tab:pretraining_domains}
\begin{tabular}{llcccccccc}
\toprule
\multirow{2}{*}{\textbf{Pretraining domain}} & \multirow{2}{*}{\textbf{Strategy}} & \multicolumn{2}{c}{\textbf{Potsdam}} & \multicolumn{2}{c}{\textbf{Vaihingen}} & \multicolumn{2}{c}{\textbf{MMHunan}} & \multicolumn{2}{c}{\textbf{GeoNRW}} \\
\cmidrule(lr){3-4}\cmidrule(lr){5-6}\cmidrule(lr){7-8}\cmidrule(lr){9-10}
 &  & \textbf{mF1} & \textbf{mIoU} & \textbf{mF1} & \textbf{mIoU} & \textbf{mF1} & \textbf{mIoU} & \textbf{mF1} & \textbf{mIoU} \\
\midrule
\multirow{4}{*}{ImageNet-RS90} & RGB-only & 88.84 & 80.37 & 83.48 & 72.49 & 53.70 & 41.39 & 63.34 & 49.01 \\
 & ExaPair & 88.97 & 80.57 & 84.52 & 74.13 & 54.28 & 41.74 & 63.67 & 49.48 \\
 & MixPair & \textbf{89.05} & \textbf{80.68} & \textbf{84.95} & \textbf{74.52} & \textbf{55.46} & \textbf{42.96} & 63.88 & 49.80 \\
 & SemPair & 89.03 & 80.64 & 84.49 & 73.94 & 54.69 & 42.41 & \textbf{64.43} & \textbf{50.06} \\
\midrule
\multirow{4}{*}{NWPU-RGBD-45} & RGB-only & 86.20 & 76.36 & 79.09 & 67.52 & 49.51 & 38.15 & 59.08 & 45.07 \\
 & ExaPair & 87.04 & 77.61 & \textbf{83.66} & \textbf{73.12} & 52.82 & \textbf{41.05} & 61.02 & 47.08 \\
 & MixPair & 87.16 & 77.80 & 82.88 & 71.99 & \textbf{53.83} & 40.91 & \textbf{61.95} & \textbf{47.96} \\
 & SemPair & \textbf{87.21} & \textbf{77.87} & 83.04 & 72.11 & 52.11 & 40.20 & 61.29 & 47.38 \\
\bottomrule
\end{tabular}
\end{table*}

As shown in Table~\ref{tab:pretraining_domains}, changing the pretraining domain affects the relative performance of the correspondence strategies, but does not alter the overall findings. With NWPU-RGBD-45 pretraining, SemPair achieves $77.87\%$ mIoU on Potsdam, slightly outperforming ExaPair at $77.61\%$. In contrast, ExaPair performs better on Vaihingen, reaching $73.12\%$ compared with $72.11\%$ for SemPair. Similar variations are observed on the other two benchmarks. For MMHunan, ExaPair, MixPair, and SemPair achieve $41.05\%$, $40.91\%$, and $40.20\%$ mIoU, respectively, all exceeding the RGB-only baseline of $38.15\%$. GeoNRW exhibits a different ranking, with MixPair achieving $47.96\%$, followed by SemPair at $47.38\%$ and ExaPair at $47.08\%$. 

Across the four downstream benchmarks, all three multimodal strategies outperform RGB-only pretraining with the alternative source dataset, while SemPair remains competitive with ExaPair. The varying rankings suggest that the effect of correspondence granularity depends on the source-target setting.

\subsection{Transfer under Different Downstream Conditions}\label{sec:transfer_conditions}
We next examine conditions that control how strongly the pretrained representation influences the final model. Two factors are varied: the amount of labeled downstream data and the degree to which the encoder is allowed to adapt.

\subsubsection{Label-Efficient Transfer}
\begin{table*}[!t]\centering
\caption{Label-efficient transfer measured by mF1 and mIoU (\%). P1, P3, and P6 denote 1, 3, and 6 labeled Potsdam training tiles. V3 and V6 denote 3 and 6 labeled Vaihingen training tiles. Full denotes the complete training split.}\label{tab:label_efficient}
\begin{tabular}{lcccccccccccccc}
\toprule
\multirow{3}{*}{\textbf{Initialization}} & \multicolumn{8}{c}{\textbf{Potsdam}} & \multicolumn{6}{c}{\textbf{Vaihingen}} \\
\cmidrule(lr){2-9}\cmidrule(lr){10-15}
 & \multicolumn{2}{c}{\textbf{P1}} & \multicolumn{2}{c}{\textbf{P3}} & \multicolumn{2}{c}{\textbf{P6}} & \multicolumn{2}{c}{\textbf{Full}} & \multicolumn{2}{c}{\textbf{V3}} & \multicolumn{2}{c}{\textbf{V6}} & \multicolumn{2}{c}{\textbf{Full}} \\
\cmidrule(lr){2-3}\cmidrule(lr){4-5}\cmidrule(lr){6-7}\cmidrule(lr){8-9}\cmidrule(lr){10-11}\cmidrule(lr){12-13}\cmidrule(lr){14-15}
 & \textbf{mF1} & \textbf{mIoU} & \textbf{mF1} & \textbf{mIoU} & \textbf{mF1} & \textbf{mIoU} & \textbf{mF1} & \textbf{mIoU} & \textbf{mF1} & \textbf{mIoU} & \textbf{mF1} & \textbf{mIoU} & \textbf{mF1} & \textbf{mIoU} \\
\midrule
Scratch & 70.91 & 56.85 & 72.83 & 58.82 & 75.60 & 62.03 & 76.83 & 63.74 & 67.84 & 54.85 & 71.23 & 58.48 & 72.19 & 59.83 \\
RGB-only & 82.89 & 71.33 & 85.07 & 74.46 & 86.52 & 76.69 & 88.84 & 80.37 & 79.44 & 67.04 & 82.67 & 71.51 & 83.48 & 72.49 \\
ExaPair & 83.42 & 72.15 & \textbf{85.54} & \textbf{75.24} & 86.86 & 77.24 & 88.97 & 80.57 & 79.89 & 67.67 & 82.83 & 71.53 & 84.52 & 74.13 \\
MixPair & \textbf{83.66} & \textbf{72.53} & 85.32 & 74.90 & \textbf{87.19} & \textbf{77.75} & \textbf{89.05} & \textbf{80.68} & 80.29 & 68.12 & 82.96 & 71.80 & \textbf{84.95} & \textbf{74.52} \\
SemPair & 83.09 & 71.64 & 85.20 & 74.71 & 87.12 & 77.61 & 89.03 & 80.64 & \textbf{80.49} & \textbf{68.43} & \textbf{83.56} & \textbf{72.62} & 84.49 & 73.94 \\
\bottomrule
\end{tabular}
\end{table*}

\begin{figure}[!t]
\centering
\includegraphics[width=\linewidth]{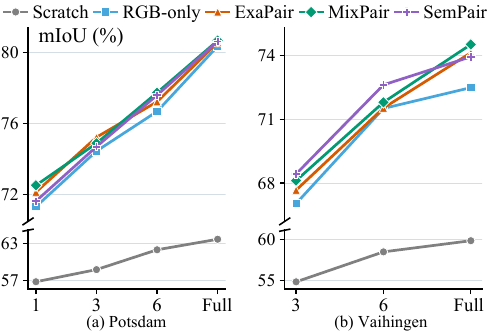}
\caption{Label-efficient transfer on Potsdam and Vaihingen.}
\label{fig:label_efficiency}
\end{figure}

The results of label-efficient transfer are summarized in Table~\ref{tab:label_efficient}. Pretraining provides substantial benefits under limited downstream supervision. With only one labeled training image from Potsdam, Scratch achieves $56.85\%$ mIoU, whereas RGB-only pretraining improves the result to $71.33\%$. Multimodal pretraining brings further gains, with ExaPair, MixPair, and SemPair achieving $72.15\%$, $72.53\%$, and $71.64\%$, respectively. A similar pattern emerges on Vaihingen with three labeled tiles, where SemPair achieves $68.43\%$ mIoU, substantially outperforming Scratch at $54.85\%$ and slightly exceeding ExaPair at $67.67\%$. These results indicate that SemPair preserves the benefits of multimodal pretraining even when downstream annotations are severely limited.

The ranking varies with the label budget, arguing against a universal preference for exact pairs. As illustrated in Figure~\ref{fig:label_efficiency}, the performance trends remain broadly comparable across the three multimodal strategies, although their relative rankings change with the amount of supervision. MixPair performs best on Potsdam with one and six training tiles and with the full split, while ExaPair is slightly better with three tiles. For Vaihingen, SemPair performs best in the low-label settings, exceeding ExaPair by $0.76\%$ with three tiles and $1.09\%$ with six. With full supervision, MixPair achieves the highest mIoU. Overall, SemPair remains competitive across different label budgets despite the absence of exact RGB-depth pairs during pretraining, indicating that effective transfer under pairing relaxation does not depend on abundant downstream annotations.

\subsubsection{Effect of Downstream Adaptation Strength}
\begin{table}[!t]\centering\scriptsize
\caption{Effect of downstream adaptation strength on mF1 and mIoU (\%). The same Scratch baseline is used across all three adaptation settings, with the entire network trained from random initialization at an initial learning rate of $10^{-2}$.}\label{tab:adaptation_strength}
\begin{tabular}{llcccccc}
\toprule
\multirow{2}{*}{\textbf{Dataset}} & \multirow{2}{*}{\textbf{Init.}} & \multicolumn{2}{c}{\textbf{Frozen}} & \multicolumn{2}{c}{\textbf{Fine-tuning}} & \multicolumn{2}{c}{\textbf{FT($10^{-2}$)}} \\
\cmidrule(lr){3-4}\cmidrule(lr){5-6}\cmidrule(lr){7-8}
 &  & \textbf{mF1} & \textbf{mIoU} & \textbf{mF1} & \textbf{mIoU} & \textbf{mF1} & \textbf{mIoU} \\
\midrule
\multirow{3}{*}{Potsdam} & Scratch & 76.83 & 63.74 & 76.83 & 63.74 & 76.83 & 63.74 \\
 & ExaPair & 86.58 & 76.93 & 88.97 & 80.57 & \textbf{89.65} & \textbf{81.61} \\
 & SemPair & \textbf{86.67} & \textbf{77.06} & \textbf{89.03} & \textbf{80.64} & 89.56 & 81.49 \\
\midrule
\multirow{3}{*}{Vaihingen} & Scratch & 72.19 & 59.83 & 72.19 & 59.83 & 72.19 & 59.83 \\
 & ExaPair & 81.93 & 70.16 & \textbf{84.52} & \textbf{74.13} & \textbf{86.01} & \textbf{76.29} \\
 & SemPair & \textbf{82.22} & \textbf{70.61} & 84.49 & 73.94 & 85.98 & 76.15 \\
\bottomrule
\end{tabular}
\end{table}

Table~\ref{tab:adaptation_strength} examines the effect of downstream adaptation strength, comparing frozen-encoder transfer with fine-tuning at encoder learning rates of $10^{-3}$ and $10^{-2}$. For Potsdam, SemPair slightly outperforms ExaPair under frozen transfer and conservative fine-tuning, by $0.13\%$ and $0.07\%$, respectively, while trailing by only $0.12\%$ under stronger adaptation. A similar pattern emerges on Vaihingen, where SemPair leads by $0.45\%$ with a frozen encoder, whereas ExaPair holds small advantages of $0.19\%$ and $0.14\%$ under the two fine-tuning settings. Overall, SemPair remains competitive across different adaptation strengths, with no consistent advantage for exact instance correspondence. The smaller performance gaps after fine-tuning suggest that downstream adaptation can partially mitigate the influence of correspondence granularity during pretraining.

\subsection{Generalization Across Model Scales}\label{sec:model_scale_generalization}

\begin{table*}[!t]\centering
\caption{Generalization across model scales measured by mF1 and mIoU (\%). The best result for each metric within each dataset and model scale is highlighted in bold.}\label{tab:model_scale}
\begin{tabular}{llcccccccccccc}
\toprule
\multirow{3}{*}{\textbf{Architecture}} &\multirow{3}{*}{\textbf{Params.}} &\multicolumn{6}{c}{\textbf{Potsdam}} &\multicolumn{6}{c}{\textbf{Vaihingen}} \\
\cmidrule(lr){3-8}\cmidrule(lr){9-14}& & \multicolumn{2}{c}{\textbf{ExaPair}}& \multicolumn{2}{c}{\textbf{MixPair}}& \multicolumn{2}{c}{\textbf{SemPair}}& \multicolumn{2}{c}{\textbf{ExaPair}}& \multicolumn{2}{c}{\textbf{MixPair}}& \multicolumn{2}{c}{\textbf{SemPair}} \\
\cmidrule(lr){3-4}\cmidrule(lr){5-6}\cmidrule(lr){7-8}\cmidrule(lr){9-10}\cmidrule(lr){11-12}\cmidrule(lr){13-14}& & \textbf{mF1} & \textbf{mIoU}& \textbf{mF1} & \textbf{mIoU}& \textbf{mF1} & \textbf{mIoU}& \textbf{mF1} & \textbf{mIoU}& \textbf{mF1} & \textbf{mIoU}& \textbf{mF1} & \textbf{mIoU} \\
\midrule
DFormer-Tiny & 5.64M& 88.91 & 80.47& \textbf{89.09} & \textbf{80.77}& 89.06 & 80.71& \textbf{83.99} & \textbf{73.44}& 83.60 & 72.66& 83.97 & 73.20 \\
DFormer-Small & 19.05M& 88.97 & 80.57& \textbf{89.05} & \textbf{80.68}& 89.03 & 80.64& 84.52 & 74.13& \textbf{84.95} & \textbf{74.52}& 84.49 & 73.94 \\
DFormer-Base & 29.95M& 89.25 & 81.02& 89.16 & 80.86& \textbf{89.30} & \textbf{81.08}& 84.72 & 74.26& \textbf{85.42} &\textbf{75.34}& 85.37 & 75.26 \\
\bottomrule
\end{tabular}
\end{table*}

Finally, we evaluate DFormer-Tiny, DFormer-Small, and DFormer-Base under the same standard fine-tuning protocol. As presented in Table~\ref{tab:model_scale}, no single pairing strategy consistently dominates across model scales. For Potsdam, both MixPair and SemPair outperform ExaPair with DFormer-Tiny, while all three strategies achieve comparable performance with DFormer-Small. SemPair achieves the highest mIoU of $81.08\%$ with DFormer-Base. A different pattern emerges on Vaihingen, where ExaPair performs best with DFormer-Tiny, whereas MixPair leads with DFormer-Small and DFormer-Base. With DFormer-Base, MixPair and SemPair achieve $75.34\%$ and $75.26\%$ mIoU, respectively, compared with $74.26\%$ for ExaPair. These results extend the evidence for correspondence relaxation beyond the default DFormer-Small encoder, covering model capacities from $5.64$M to $29.95$M parameters.

\subsection{Limitations and Discussion}\label{sec:limitations_discussion}
Our results show that exact instance correspondence is not a prerequisite for effective multimodal pretraining in the remote-sensing settings considered in this work. Nevertheless, the present study is based on controlled re-pairing: the original RGB-pseudo-depth correspondence is known, while SemPair deliberately replaces exact pairs with class-consistent cross-instance pairs during pretraining. This design allows correspondence granularity to be isolated from other factors, but it does not yet establish the same conclusion for independently collected multimodal archives in real scenes.

SemPair also assumes that semantic consistency provides a meaningful cross-modal relation. This assumption is well aligned with semantic recognition and segmentation, but exact instance correspondence may remain important for tasks that intrinsically depend on local geometric or temporal consistency, such as registration, reconstruction, or fine-grained change localization. SemPair should therefore be viewed as a relaxed alternative to ExaPair rather than a universal replacement for paired supervision.

Although all downstream experiments in this work focus on remote sensing, the formulation itself is not tied to a specific sensor combination or backbone. Its essential requirement is a reliable semantic relation across modalities rather than exact instance alignment. This makes the same principle potentially applicable to other multimodal settings where large-scale paired data are difficult to obtain. Such generalization, however, remains to be empirically validated beyond the remote-sensing scenarios studied here.

\section{Conclusion}\label{sec:conclu}
This work challenges the assumption that effective multimodal pretraining requires exact instance correspondence. By introducing SemPair and MixPair, we demonstrate that relaxing cross-modal pairing can preserve transferable representations without modifying the network architecture or learning objective. The latent-factor analysis characterizes how pairing granularity changes cross-modal correlations, while extensive experiments across pretraining domains, downstream benchmarks, and adaptation settings show that SemPair remains competitive with ExaPair. These findings suggest that exact instance correspondence is not a prerequisite for effective multimodal remote sensing pretraining, and that semantic-level correspondence provides a viable alternative. Future work will extend this controlled study to independently collected multimodal archives with incomplete cross-modal correspondence.

\bibliographystyle{IEEEtran}
\bibliography{references}
\end{document}